\documentclass[letterpaper, 10 pt, conference]{ieeeconf}  

\usepackage{epsfig}
\usepackage{graphicx}
\usepackage{amsmath}
\usepackage{amssymb}

\usepackage{algorithm}
\usepackage{algorithmic}
\usepackage{multirow}

\usepackage{colortbl}
\usepackage{xcolor}

\usepackage{subfigure}
\usepackage{cite}

\usepackage{hyperref}

\def \NUM {eleven }

\IEEEoverridecommandlockouts                              

\title{\LARGE \bf
Planar Object Tracking in the Wild: A Benchmark
}

\author{Pengpeng Liang$^{1,3}$, Yifan Wu$^{2}$, Hu Lu$^{4}$, Liming Wang$^{1}$, Chunyuan Liao$^{3}$, Haibin Ling$^{2,3*}$
\thanks{$^{1}$Pengpeng Liang and Liming Wang are with School of Information Engineering, Zhengzhou University, Zhengzhou 450001, China.
        {\{ieppliang, ielmwang\}@zzu.edu.cn}}%
\thanks{$^{2}$Yifan Wu and Haibin Ling are with Computer \& Information Sciences Department, Temple University, Philadelphia, USA.
        \{yifan.wu, hbling\}@temple.edu}
\thanks{$^{3}$HiScene Information Technologies, Shanghai 201210, China.}
\thanks{$^{4}$Hu Lu is with School of Computer Science and Communication Engineering, Jiangsu University, Zhenjiang 212003, China.  luhu@ujs.edu.cn}
\thanks{*Correspondence author.}
}

\begin{document}

\maketitle
\thispagestyle{empty}
\pagestyle{empty}

\begin{abstract}
Planar object tracking is an actively studied problem in vision-based robotic applications. While several benchmarks have been constructed for evaluating state-of-the-art algorithms, there is a lack of video sequences captured in the wild rather than in constrained laboratory environment. In this paper, we present a carefully designed planar object tracking benchmark containing 210 videos of 30 planar objects sampled in the natural environment. In particular, for each object, we shoot seven videos involving various challenging factors, namely \emph{scale change}, \emph{rotation}, \emph{perspective distortion}, \emph{motion blur}, \emph{occlusion}, \emph{out-of-view}, and \emph{unconstrained}. The ground truth is carefully annotated semi-manually to ensure the quality. Moreover, \NUM state-of-the-art algorithms are evaluated on the benchmark using two evaluation metrics, with detailed analysis provided for the evaluation results. We expect the proposed benchmark to benefit future studies on planar object tracking.
\end{abstract}

\section{Introduction}
\label{sec:introduction}
Camera localization and environment modeling is a fundamental problem in vision-based robotics. In theory, these tasks can be completed by tracking and then analyzing 3D structures in the input from visual sensors. In practice, however, tracking of 3D structures is by itself very challenging. Two-dimensional planar structures, instead, often serve as a reliable and reasonable surrogate. As a result, planar object tracking plays an important role in many vision-based robotic applications, such as visual servoing~\cite{hutchinson1996tutorial}, visual SLAM~\cite{concha2015dpptam}, and UAV control~\cite{mondragon20103d},  as well as related fields, e.g. augmented reality~\cite{klein2007parallel,kato1999marker}.

\begin{figure}[!t]
\subfigure[The Metaio dataset \cite{lieberknecht2009dataset}]{
\includegraphics[width=0.475\linewidth,height=0.3\linewidth]{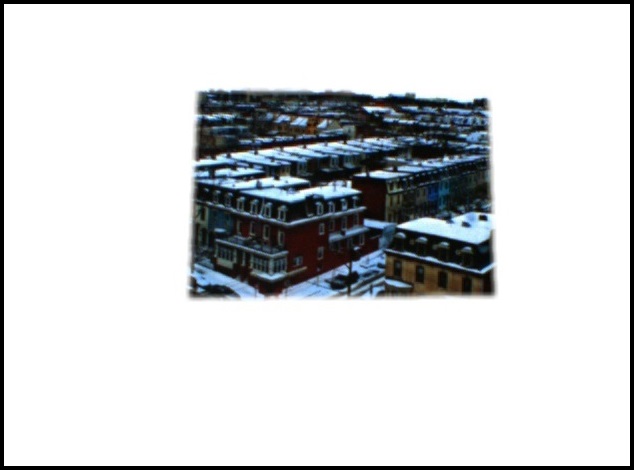}
}
\hspace{-8pt}
\subfigure[The TMT dataset \cite{roy2015tracking}]{
\includegraphics[width=0.475\linewidth,height=0.3\linewidth]{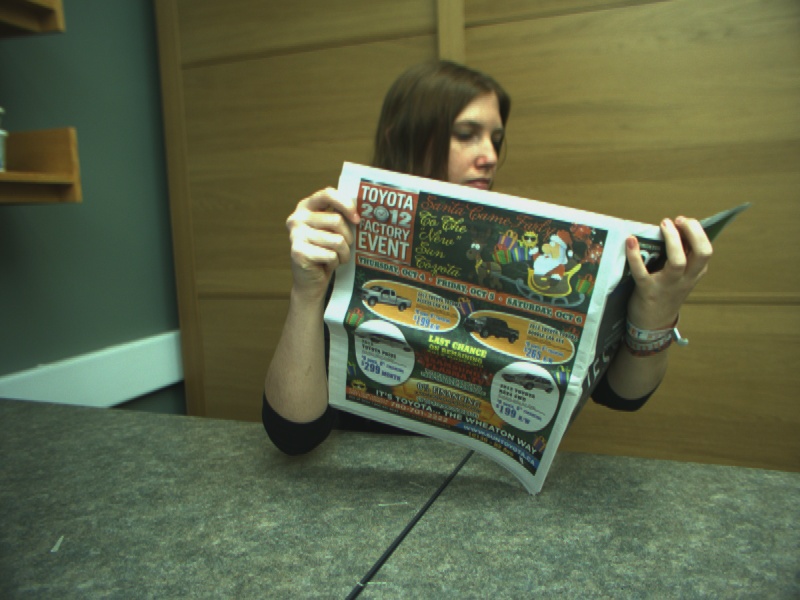}
}
\vspace{-2mm}\\
\subfigure[The planar texture dataset  \cite{gauglitz2011evaluation}]{
\includegraphics[width=0.475\linewidth,height=0.3\linewidth]{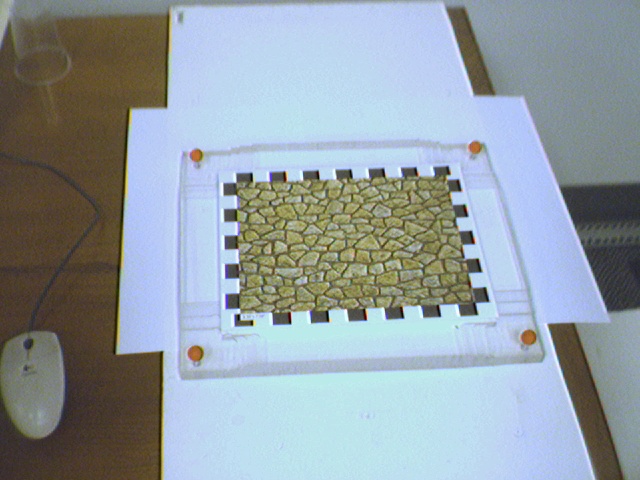}
}
\hspace{-8pt}
\subfigure[The proposed benchmark]{
\includegraphics[width=0.475\linewidth,height=0.3\linewidth]{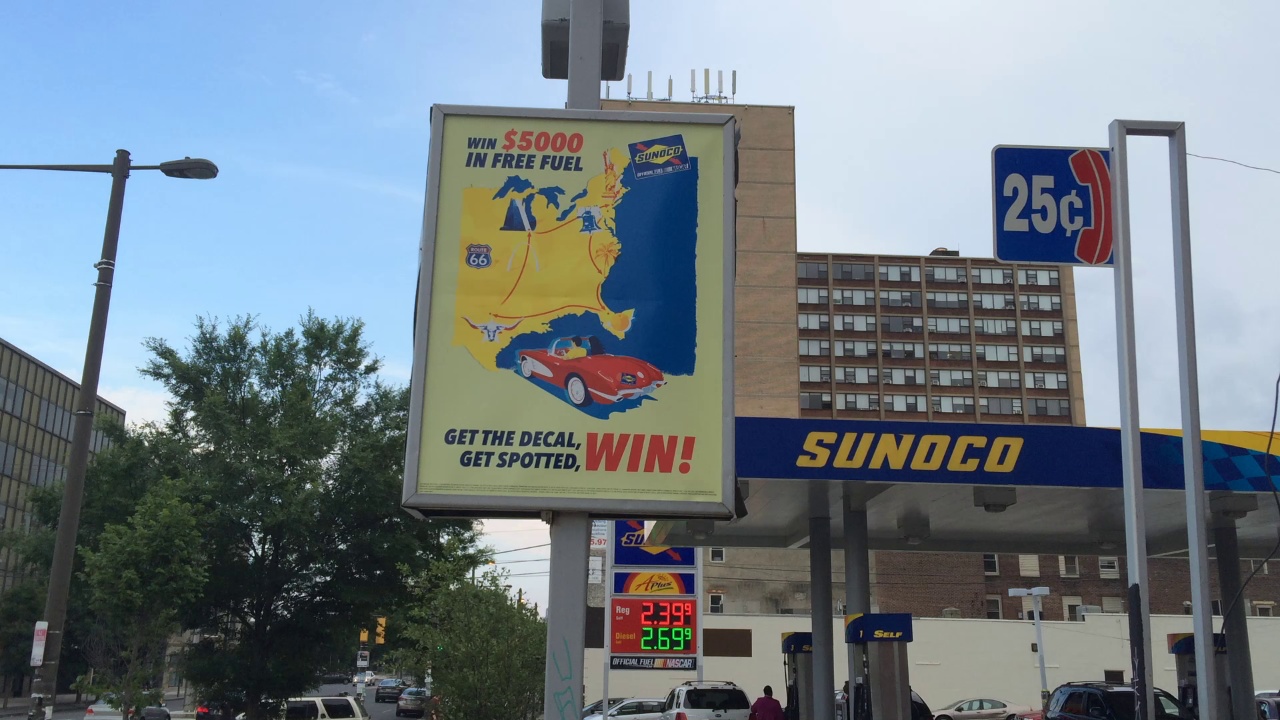}
}
\caption{Sample frames from three representative benchmarks and ours.  Note: frames in the Metaio dataset have artificial white background by design, and we draw the image boundary for better illustration. }
\label{fig:comparsion}
\end{figure}

Recently, several datasets have been provided for comprehensively evaluating planar tracking, including the Metaio dataset \cite{lieberknecht2009dataset}, the tracking manipulation tasks (TMT) dataset \cite{roy2015tracking} and the planar texture dataset \cite{gauglitz2011evaluation}. Though these datasets overcome the shortcomings of synthetic datasets that cannot faithfully reproduce the real effects of every condition, all of them are constructed in laboratory environments (see Fig.~\ref{fig:comparsion}). A disadvantage of the datasets collected this way is that the background is short of diversity or even artificial, while in real world scenarios can be much more complicated. Consequently, it is insufficient to evaluate the effectiveness of planar object tracking algorithms in natural setting with these datasets.

To address the above issue, in this paper, we present a novel planar object tracking benchmark containing 210 video sequences collected \emph{in the wild} and each sequence has 500 frames plus an additional frame for initialization. For constructing the dataset, we first select 30 planar objects in natural scene; then, for each object, we capture seven videos involving seven challenging factors. Six of the challenging factors are commonly encountered in practical applications, while the seventh dedicates to an unconstrained condition, typically involving multiple challenging factors. To annotate the ground truth as precisely as possible, given the initial state of an object, we first run a keypoint-based object tracking algorithm using structured output learning \cite{hare2012efficient} as an initial guess; then we manually check and revise the results to ensure accuracy, with tracking re-initialization if needed. We annotate every other frame for each sequence.

To understand the performance of state-of-the-arts, we evaluate \NUM  modern tracking algorithms on the dataset. These algorithms include three types of trackers: four keypoint-based planar object tracking algorithms \cite{hare2012efficient,ozuysal2010fast,lowe2004distinctive,bay2008speeded}, four region-based (a.k.a.\ direct methods) planar object tracking algorithms \cite{richa2011visual, benhimane2004real,baker2004lucas,goesm2017}, and three generic object tracking algorithms \cite{kwon2014geometric,bao2012real,ross2008incremental}. We use two performance metrics to analyze the evaluation results in details. One metric is based on four reference points and measures the distance of misalignment between the ground truth state and the predicted state; the other is the difference between the ground truth homography and the predicted homography. Note that we do not evaluate the state-of-the-art generic object trackers such as trackers using deep learned features~\cite{nam2016learning,qi2016hedged}. This is because such trackers, by outputting rectangular bounding boxes, aim at locating the target rather than providing the precise state of the target.

In summary, our contributions are three-fold: (1) collecting systematically a dataset containing 210 videos for planar object tracking in the wild; (2)  providing accurate ground truth by annotating the data in a semi-automatic manner, and 52,710 frames are annotated in total; and (3) evaluating \NUM representative state-of-the-art algorithms with two performance metrics, and analyzing the results in details according to seven different motion patterns. To the best of our knowledge, our benchmark not only is the largest one to date, but also is more realistic than previously proposed ones. 
%
The benchmark, along with the evaluation results, is made available for research purpose (\url{http://www.dabi.temple.edu/~hbling/data/POT-210/planar_benchmark.html}).

In the rest of the paper, we first summarize related work in Sec.~\ref{sec:related} and then introduce details of the dataset in Sec.~\ref{sec:dataset}. The evaluation and the analysis of the results are described in Sec.~\ref{sec:evaluation}. Finally, we conclude the paper in Sec.~\ref{sec:conclusion}.

\section{Related Work}
\label{sec:related}

\subsection{Previous benchmark}
With the advance of planar object tracking, it is crucial to provide benchmarks for evaluation purpose. Recently, there have been several such benchmarks relevant with our work \cite{lieberknecht2009dataset}, \cite{roy2015tracking} and  \cite{gauglitz2011evaluation}. In \cite{lieberknecht2009dataset}, the authors collected 40 sequences with eight different texture patterns under five different dynamic behaviors. To annotate the ground truth precisely, a camera was mounted on a robotic measurement arm which could record the camera pose. One limitation of using the measurement arm for annotation is that it may have problems when used in natural environments flexibly.

To evaluate tracking algorithms for manipulation tasks, 100 sequences were collected and each sequence was tagged with different challenging factors in~\cite{roy2015tracking}. For annotation, three trackers were used to annotate the ground truth, and the coordinates of the four reference corners were determined when the coordinates reported by all the three trackers lay within a certain range. Such annotation avoids heavy manual work, but can be noisy especially for challenging sequences on which at least one tracker fails.

In \cite{gauglitz2011evaluation}, 96 sequences were collected with six planar textures under 16 different motion patterns each. To annotate the ground truth in a semi-automatic manner, a planar texture picture was held by a milled acrylic glass frame and there were four bright red balls on the frame as markers.

Besides the above three benchmark datasets, the authors of several papers focusing on tracking algorithms collected their own data for evaluation purpose. In \cite{hare2012efficient}, five sequences were collected and the ground truth was obtained using a SLAM system which could track the 3D camera pose in each frame. In \cite{zimmermann2009tracking}, image sequences of three different objects were collected and the ground truth was annotated manually using the object corners. In \cite{SMM_tracker}, the authors used the five sequences from \cite{hare2012efficient}  and another four sequences collected by themselves to evaluate their algorithm.

It is worth mentioning that several benchmarks for generic object tracking have been proposed in recent years \cite{wu2015object,smeulders2014visual,li2016nus,kristan2015visual}. However, all of these datasets provide rectangular bounding box annotation, and none of them can be used for evaluating planar object tracking algorithms.

To the best of our knowledge, our work is the first one providing a dataset for planar object tracking in the wild. Moreover, our dataset contains 210 sequences with careful annotation, and is much larger than previous ones.

\begin{figure*}[!t]
\centering
\scriptsize
\begin{tabular}
{@{\hspace{.0mm}}c@{\hspace{.4mm}} @{\hspace{.4mm}}c@{\hspace{.4mm}}
 @{\hspace{.4mm}}c@{\hspace{.4mm}} @{\hspace{.4mm}}c@{\hspace{.4mm}}
 @{\hspace{.4mm}}c@{\hspace{.4mm}} @{\hspace{.4mm}}c@{\hspace{.0mm}}}
\includegraphics[width=0.162\linewidth,height=0.06\linewidth]{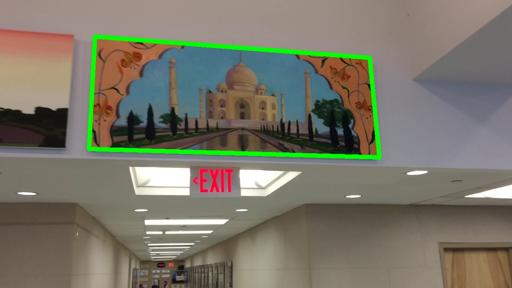}&
\includegraphics[width=0.162\linewidth,height=0.06\linewidth]{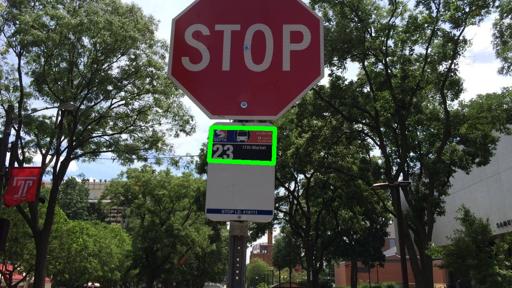}&
\includegraphics[width=0.162\linewidth,height=0.06\linewidth]{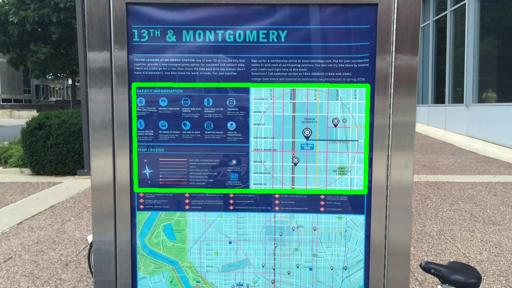}&
\includegraphics[width=0.162\linewidth,height=0.06\linewidth]{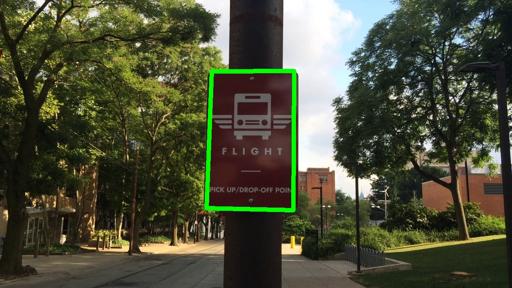}&
\includegraphics[width=0.162\linewidth,height=0.06\linewidth]{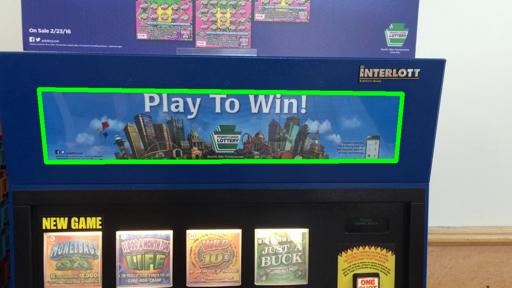}&
\includegraphics[width=0.162\linewidth,height=0.06\linewidth]{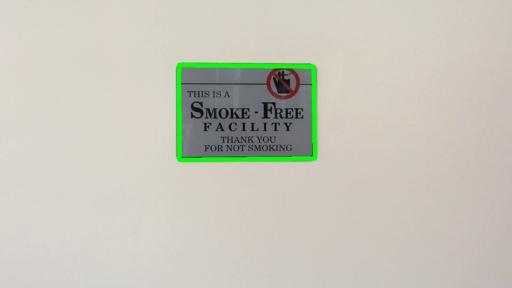}\\
\vspace{2pt}
Painting-2, 0.853&BusStop, 0.844&IndegoStation, 0.831&ShuttleStop, 0.821&Lottery-2, 0.798&SmokeFree, 0.796\\
\includegraphics[width=0.162\linewidth,height=0.06\linewidth]{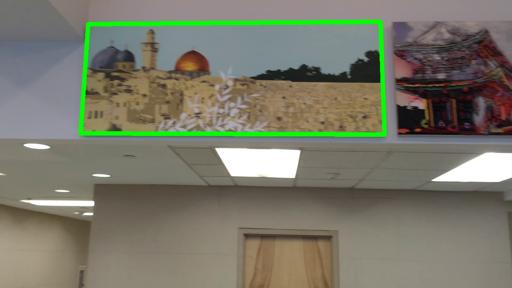}&
\includegraphics[width=0.162\linewidth,height=0.06\linewidth]{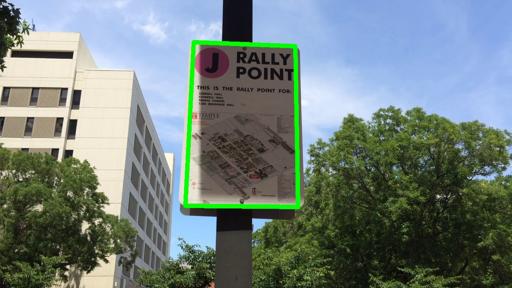}&
\includegraphics[width=0.162\linewidth,height=0.06\linewidth]{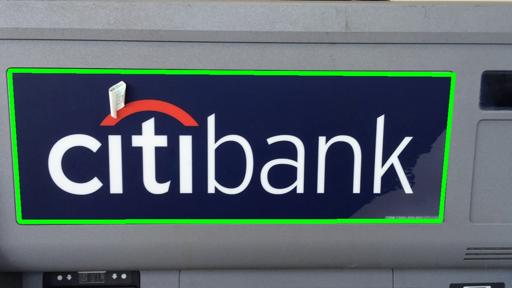}&
\includegraphics[width=0.162\linewidth,height=0.06\linewidth]{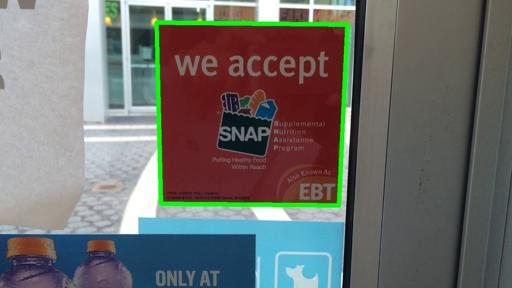}&
\includegraphics[width=0.162\linewidth,height=0.06\linewidth]{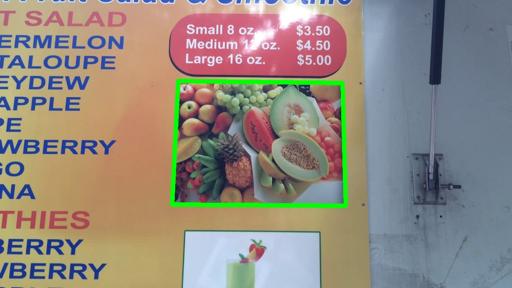}&
\includegraphics[width=0.162\linewidth,height=0.06\linewidth]{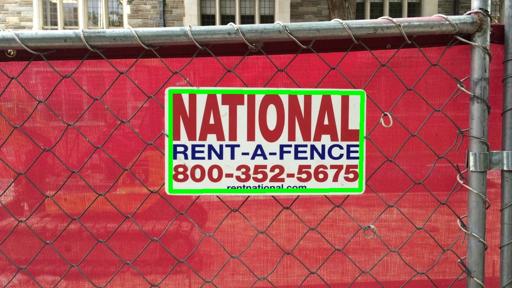}\\
\vspace{2pt}
Painting-1, 0.790&Map-1, 0.788&Citibank, 0.785&Snap, 0.760&Fruit, 0.735&Poster-2, 0.733\\
\includegraphics[width=0.162\linewidth,height=0.06\linewidth]{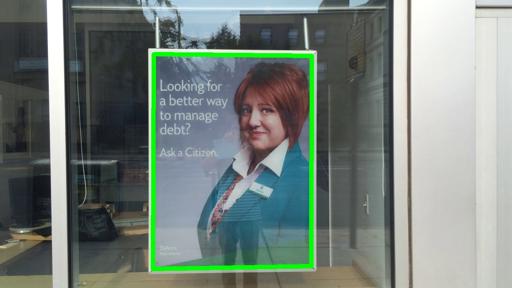}&
\includegraphics[width=0.162\linewidth,height=0.06\linewidth]{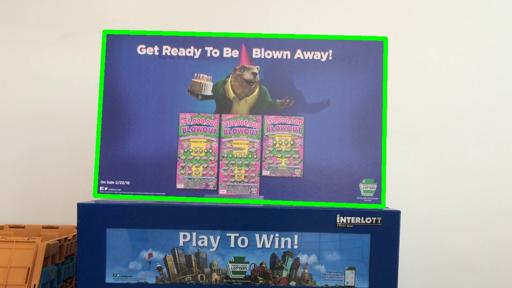}&
\includegraphics[width=0.162\linewidth,height=0.06\linewidth]{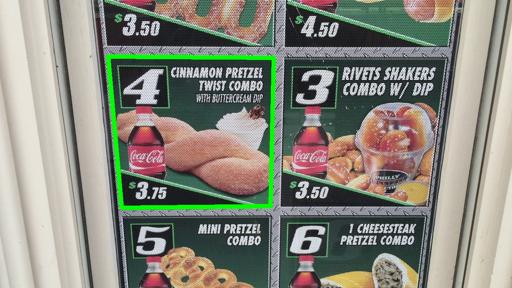}&
\includegraphics[width=0.162\linewidth,height=0.06\linewidth]{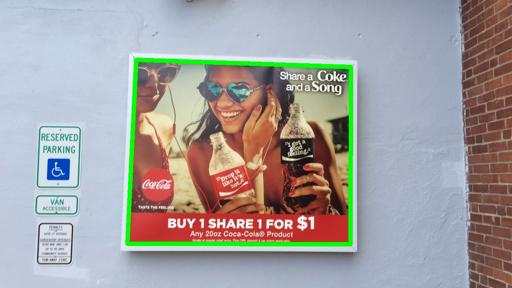}&
\includegraphics[width=0.162\linewidth,height=0.06\linewidth]{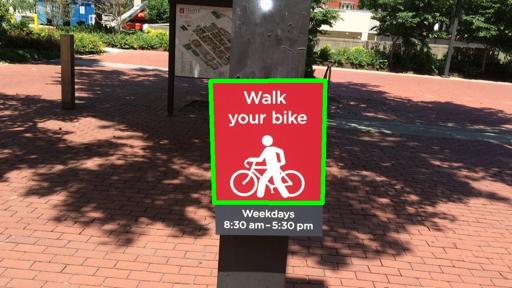}&
\includegraphics[width=0.162\linewidth,height=0.06\linewidth]{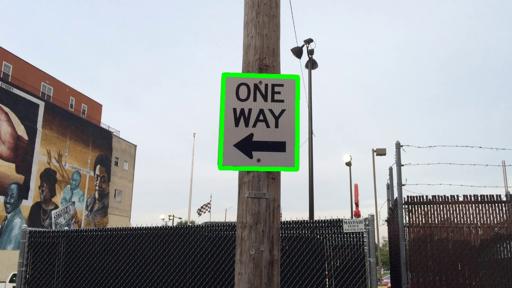}\\
\vspace{2pt}
Woman, 0.724&Lottery-1, 0.721&Pretzel, 0.721&Coke, 0.704&WalkYourBike, 0.699&OneWay, 0.697\\
\includegraphics[width=0.162\linewidth,height=0.06\linewidth]{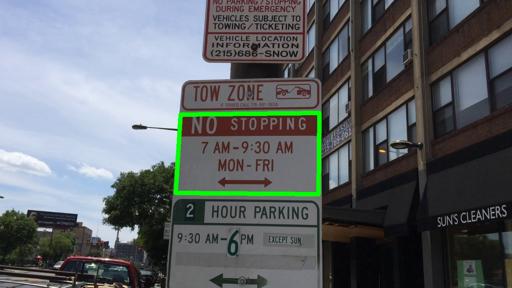}&
\includegraphics[width=0.162\linewidth,height=0.06\linewidth]{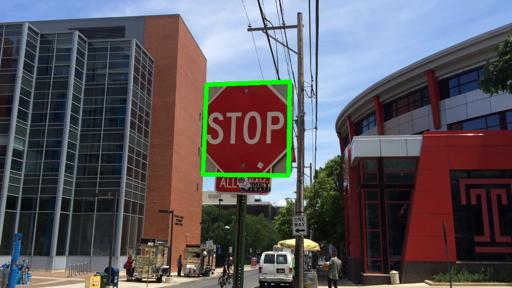}&
\includegraphics[width=0.162\linewidth,height=0.06\linewidth]{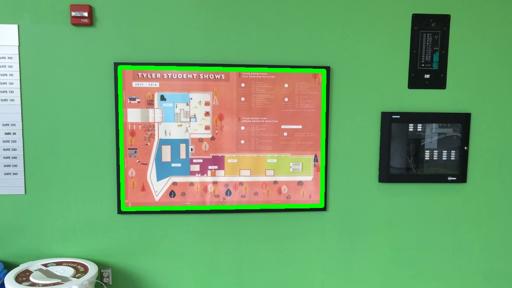}&
\includegraphics[width=0.162\linewidth,height=0.06\linewidth]{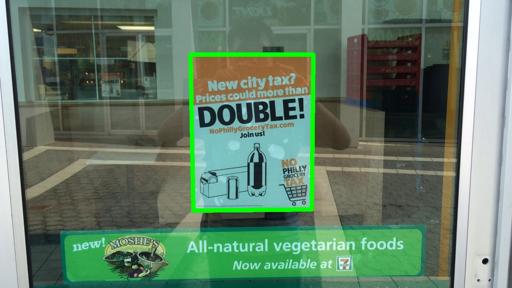}&
\includegraphics[width=0.162\linewidth,height=0.06\linewidth]{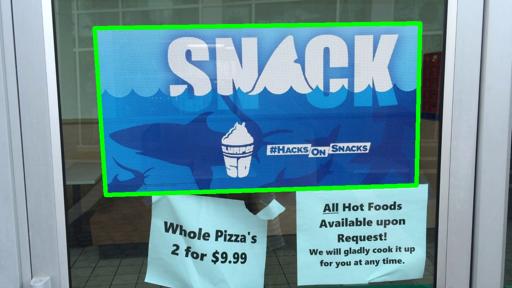}&
\includegraphics[width=0.162\linewidth,height=0.06\linewidth]{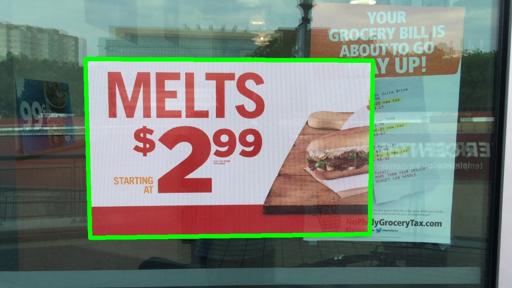}\\
\vspace{2pt}
NoStopping, 0.690&StopSign, 0.681&Map-2, 0.676&Poster-1, 0.659&Snack, 0.643&Melts, 0.640\\
\includegraphics[width=0.162\linewidth,height=0.06\linewidth]{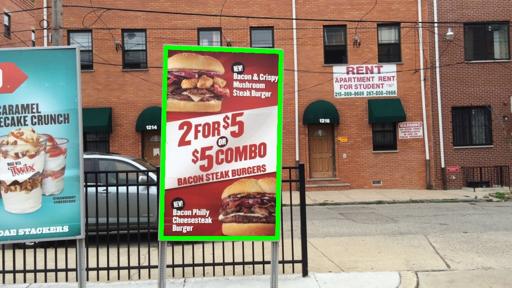}&
\includegraphics[width=0.162\linewidth,height=0.06\linewidth]{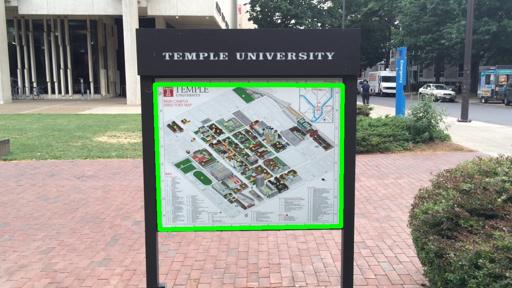}&
\includegraphics[width=0.162\linewidth,height=0.06\linewidth]{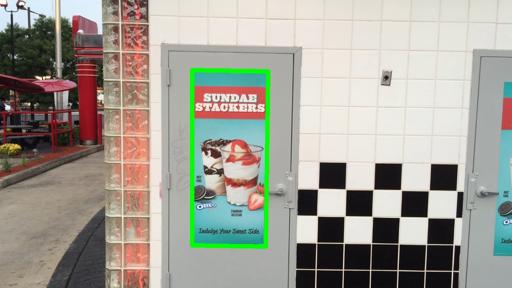}&
\includegraphics[width=0.162\linewidth,height=0.06\linewidth]{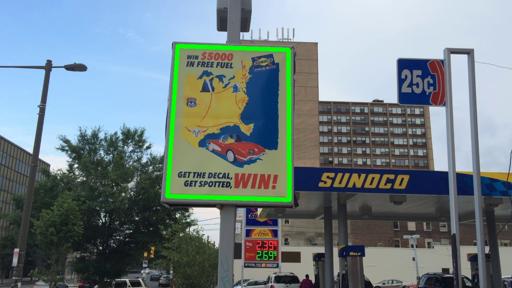}&
\includegraphics[width=0.162\linewidth,height=0.06\linewidth]{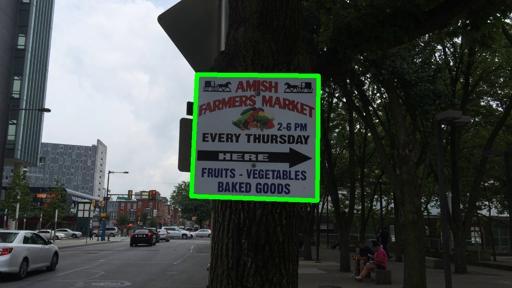}&
\includegraphics[width=0.162\linewidth,height=0.06\linewidth]{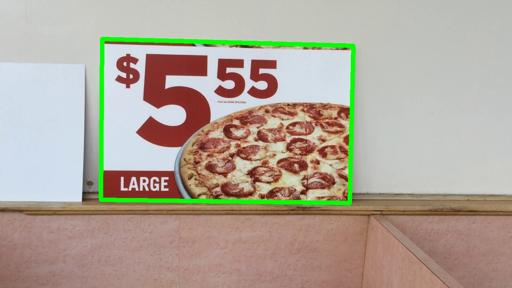}\\
\vspace{2pt}
Burger, 0.624&Map-3, 0.615&Sundae, 0.615&Sunoco, 0.595&Amish, 0.594&Pizza, 0.519\\
\end{tabular}
\caption{The 30 planar objects (in green bounding box) in our dataset, ordered from hardest to easiest according to the \emph{degree of difficulty} (Sec.~\ref{sec:metrics}).}
\label{fig:object-show}
\end{figure*}

\subsection{Tracking algorithms}
Current planar object tracking algorithms can be categorized into two main groups. The first group is keypoint-based. The algorithms \cite{hare2012efficient,lepetit2006keypoint,SMM_tracker,wagner2010real,ozuysal2010fast} lying in this group often model an object with a set of keypoints (e.g., SIFT \cite{lowe2004distinctive}, SURF \cite{bay2008speeded} and FAST \cite{rosten2010faster}) and associated descriptors, and the tracking process consists of two steps. First, a set of correspondences between object and image keypoints is constructed through descriptor matching; then, the transformation of the object in the image is estimated using a robust geometric estimation algorithm such as RANSAC \cite{fischler1981random} based on the correspondences. In \cite{lepetit2006keypoint}, keypoint matching was formulated as a multi-class classification problem so that the computational burden was shifted to the training phase. In \cite{SMM_tracker}, to utilize the temporal and spatial consistency during the tracking process, a robust keypoint-based appearance model was learned with a metric learning driven approach. The authors of \cite{wagner2010real} carefully modified the feature descriptors SIFT \cite{lowe2004distinctive} and Ferns \cite{ozuysal2007fast} so that they could work at real-time speed on mobile phones. Graph matching is integrated for matching keypoints in~\cite{Wang&Ling18pami} recently.

The second group of planar tracking algorithms are region-based and sometimes called \emph{direct methods}. These algorithms \cite{richa2011visual,benhimane2004real,pressigout2005real,ito2011accurate,baker2004lucas,tan2014multi,goesm2017} lying in this group directly estimate the transformation parameters by minimizing an error that measures the image similarity between the template and its projection in the image. In \cite{pressigout2005real}, both texture and contour information were used to construct the appearance model, and the 2D transformation was estimated by minimizing the error between the multi-cue template and the projected image patch. To deal with resolution degradation, the authors in \cite{ito2011accurate} proposed to reconstruct the target model with an image sampling process. In \cite{tan2014multi}, random forest was used to learn the relationship between the parameters modeling the motion of the target and the image intensity change of the template. This learning-based approach is useful to avoid local minimum and handle partial occlusion. The authors of \cite{singh2016modular} provided a code framework for region-based trackers, also known as registration based tracking or direct visual tracking, by decomposing this kind of trackers into three modules including an appearance model, a state space model and a search method. 


In this paper, we select four keypoint-based \cite{lowe2004distinctive,bay2008speeded,hare2012efficient,ozuysal2010fast}, four region-based \cite{richa2011visual,benhimane2004real,baker2004lucas,goesm2017} and three generic object tracking algorithms \cite{kwon2014geometric,bao2012real,ross2008incremental} as representative trackers in evaluation. The details of these algorithms are given in Sec.~\ref{sec:evaluation}.

\begin{figure*}[!t]
\subfigure[Scale change]{
\label{fig:scale}
\includegraphics[width=0.158\linewidth,height=0.06\linewidth]{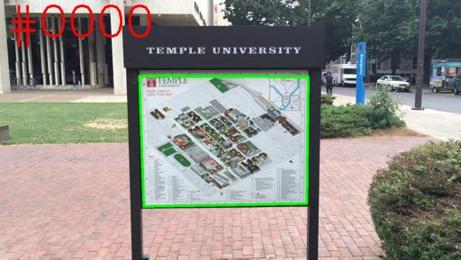}
\hspace{-3pt}
\includegraphics[width=0.158\linewidth,height=0.06\linewidth]{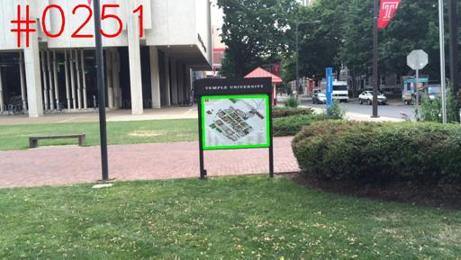}
\hspace{-3pt}
\includegraphics[width=0.158\linewidth,height=0.06\linewidth]{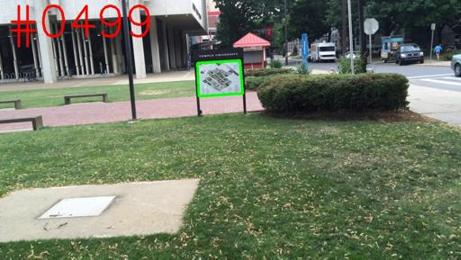}
}
\hspace{-7pt}
\subfigure[Rotation]{
\label{fig:rotation}
\includegraphics[width=0.158\linewidth,height=0.06\linewidth]{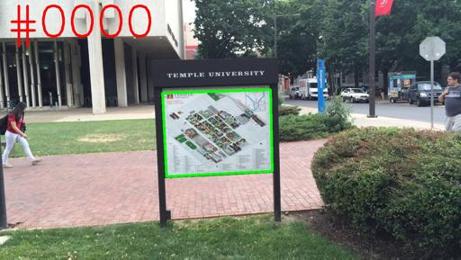}
\hspace{-3pt}
\includegraphics[width=0.158\linewidth,height=0.06\linewidth]{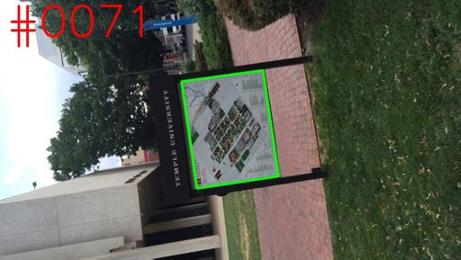}
\hspace{-3pt}
\includegraphics[width=0.158\linewidth,height=0.06\linewidth]{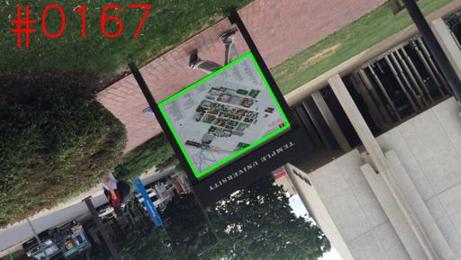}
}
\\
\vspace{-2pt}\subfigure[Perspective distortion]{
\label{fig:perspective}
\includegraphics[width=0.158\linewidth,height=0.06\linewidth]{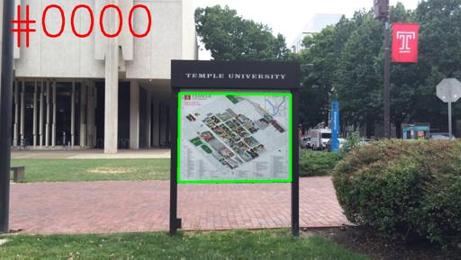}
\hspace{-3pt}
\includegraphics[width=0.158\linewidth,height=0.06\linewidth]{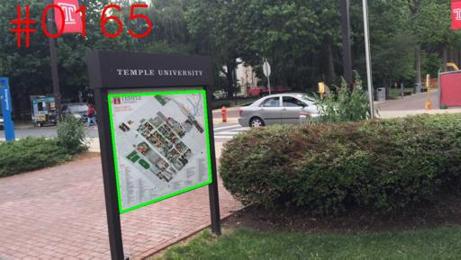}
\hspace{-3pt}
\includegraphics[width=0.158\linewidth,height=0.06\linewidth]{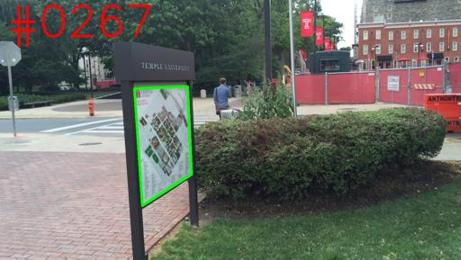}
}
\hspace{-7pt}
\subfigure[Motion blur]{
\label{fig:mb}
\includegraphics[width=0.158\linewidth,height=0.06\linewidth]{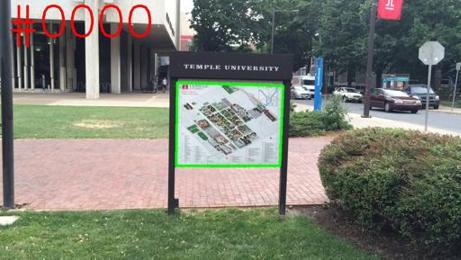}
\hspace{-3pt}
\includegraphics[width=0.158\linewidth,height=0.06\linewidth]{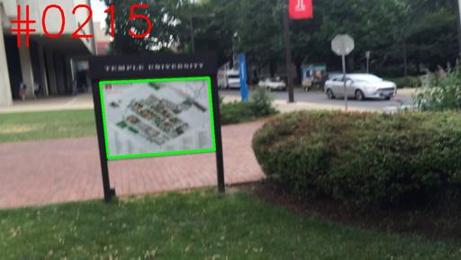}
\hspace{-3pt}
\includegraphics[width=0.158\linewidth,height=0.06\linewidth]{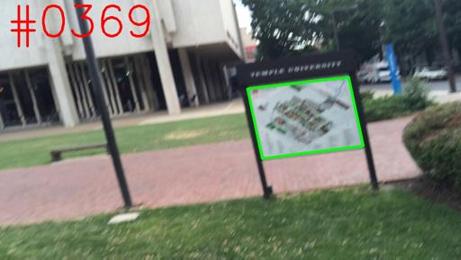}
}
\\
\vspace{-2pt}\subfigure[Occlusion]{
\label{fig:occlusion}
\includegraphics[width=0.158\linewidth,height=0.06\linewidth]{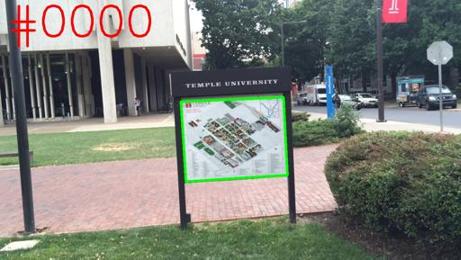}
\hspace{-3pt}
\includegraphics[width=0.158\linewidth,height=0.06\linewidth]{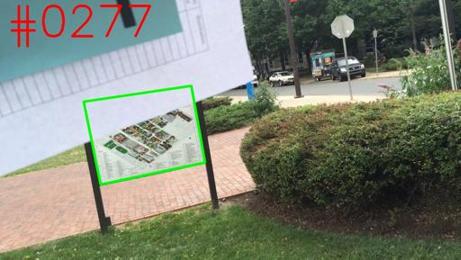}
\hspace{-3pt}
\includegraphics[width=0.158\linewidth,height=0.06\linewidth]{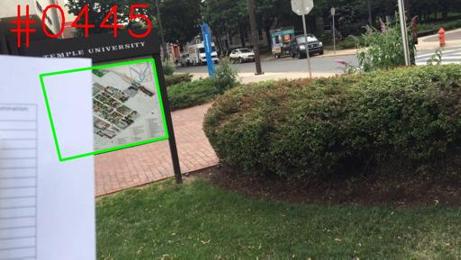}
}
\hspace{-7pt}
\subfigure[Out-of-view]{
\label{fig:ov}
\includegraphics[width=0.158\linewidth,height=0.06\linewidth]{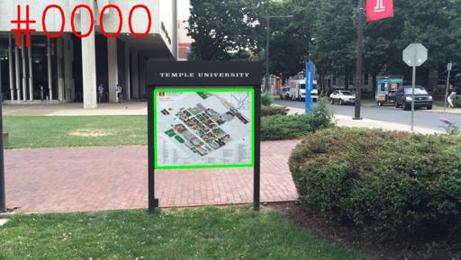}
\hspace{-3pt}
\includegraphics[width=0.158\linewidth,height=0.06\linewidth]{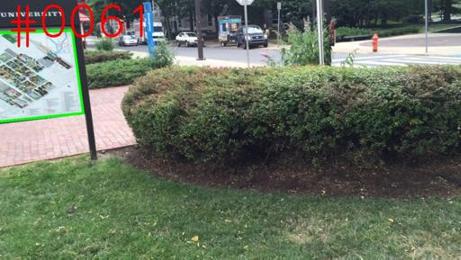}
\hspace{-3pt}
\includegraphics[width=0.158\linewidth,height=0.06\linewidth]{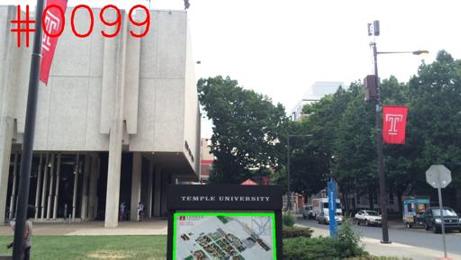}
}
\\
\vspace{-2pt}\subfigure[Unconstrained]{
\label{fig:unconstrained}
\includegraphics[width=0.158\linewidth,height=0.06\linewidth]{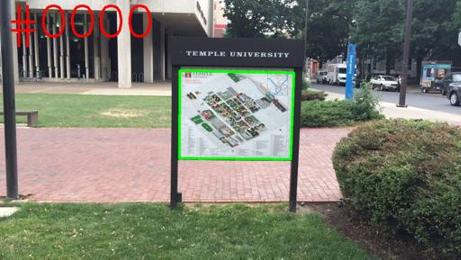}
\hspace{-3pt}
\includegraphics[width=0.158\linewidth,height=0.06\linewidth]{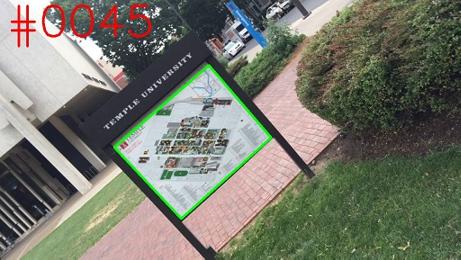}
\hspace{-3pt}
\includegraphics[width=0.158\linewidth,height=0.06\linewidth]{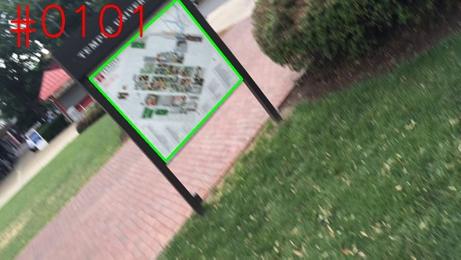}
\hspace{-3pt}
\includegraphics[width=0.158\linewidth,height=0.06\linewidth]{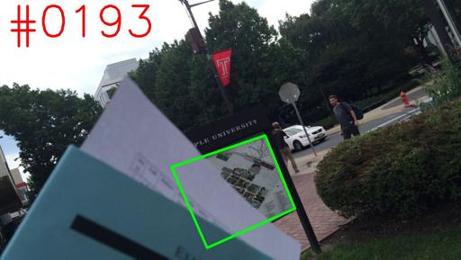}
\hspace{-3pt}
\includegraphics[width=0.158\linewidth,height=0.06\linewidth]{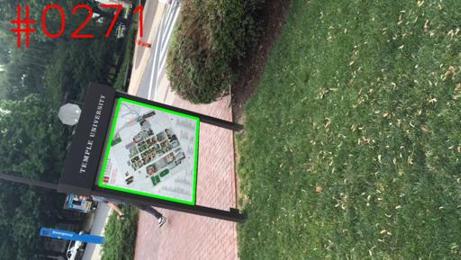}
\hspace{-3pt}
\includegraphics[width=0.158\linewidth,height=0.06\linewidth]{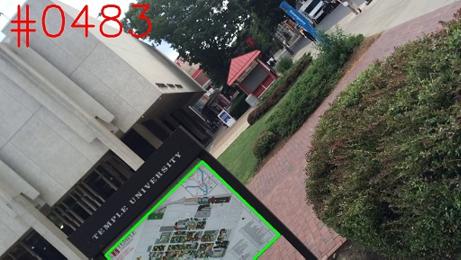}
}
\caption{Example frames for different challenging factors.}
\label{fig:motion}
\end{figure*}

\section{Dataset Design}
\label{sec:dataset}
\subsection{Dataset Construction}

We use a smart phone (iPhone 5S) to record all the videos and the camera is held by hands. The reason for using a smart phone is that it can approach everyday scenarios as closely as possible. The videos are recorded at 30 frames per second with a resolution of $1920\times 1080$, and we resample the video sequences to $1280\times 720$ for efficiency\footnote{By contrast,
the frame size in \cite{lieberknecht2009dataset} and \cite{gauglitz2011evaluation} is $640\times 480$; and the frame size in \cite{roy2015tracking} is $800\times 600$.}.

We select 30 planar objects in natural scene in different photometric environments as shown in Fig.~\ref{fig:object-show}. As we can see, the background of the selected objects varies a lot, especially when compared with previous benchmarks as shown in Fig.~\ref{fig:comparsion}.  For each object, we shoot videos involving seven motion patterns so that the dataset can be used to systematically analyze the strengths and weaknesses of different tracking algorithms. The dataset contains 210 sequences in total, and each sequence has 500 frames plus an additional frame for initialization. The following are the challenging factors involved:
\begin{itemize}
\item \textbf{Scale change (SC):} the distance between the camera and the target changes significantly (Fig.~\ref{fig:scale}).
\item \textbf{Rotation (RT):} rotating the camera  and trying to keep the camera in the same plane during rotation (Fig.~\ref{fig:rotation}).
\item \textbf{Perspective distortion (PD):}  changing the perspective between the object and the camera (Fig.~\ref{fig:perspective}).
\item \textbf{Motion blur (MB):} motion blur is generated by fast camera movement (Fig.~\ref{fig:mb}).
\item \textbf{Occlusion (OCC):} manually occluding the object while moving the camera (Fig.~\ref{fig:occlusion}).
\item \textbf{Out-of-view: (OV):}  part of the object is out of the image (Fig.~\ref{fig:ov}).
\item \textbf{Unconstrained (UC):} moving the camera freely and the resulting video sequence may involve one or more of the above challenging factors (Fig.~\ref{fig:unconstrained}).
\end{itemize}
It is worth noting that as it is hard to control the illumination condition in natural environment, illumination variation is not included in the challenging factors.

\subsection{Annotating the ground truth}

\begin{figure*}[!tp]
\centering
\subfigure[Normal]{
\label{fig:tool-normal}
\includegraphics[width=0.3\linewidth,height=0.13\linewidth]{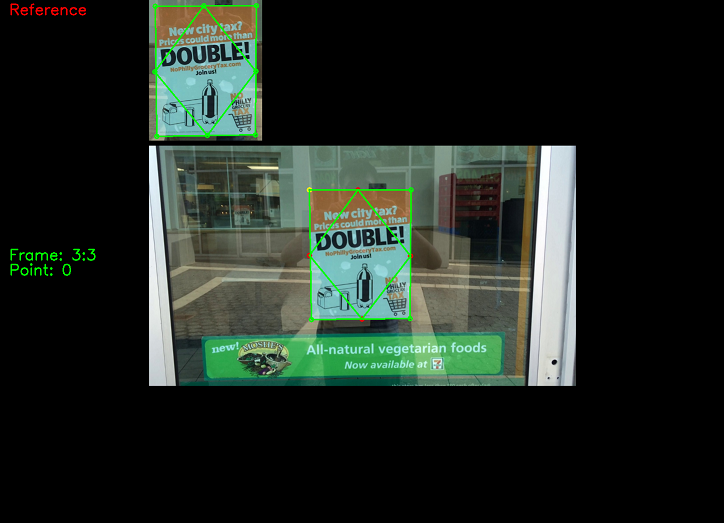}
}
\subfigure[Occlusion]{
\label{fig:tool-occ}
\includegraphics[width=0.3\linewidth,height=0.13\linewidth]{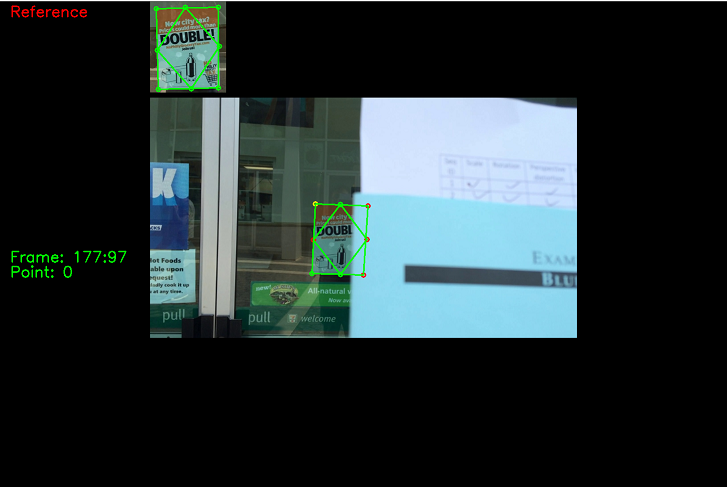}
}
\subfigure[Out-of-view]{
\label{fig:tool-ov}
\includegraphics[width=0.3\linewidth,height=0.13\linewidth]{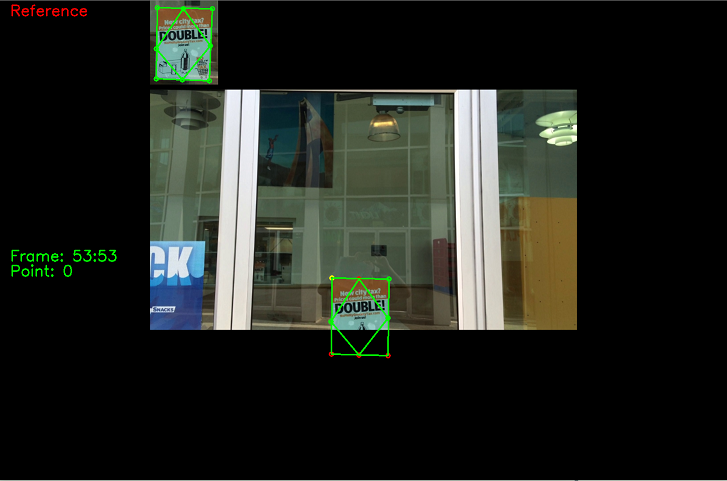}
}
\caption{The user interface of our annotation tool for different situations.}
\label{fig:annotation-tool}
\end{figure*}

Following the popular strategy in planar object tracking \cite{gauglitz2011evaluation}, we define the tracking ground truth as a transformation matrix that projects a point $p_j$ in frame $j$ to its corresponding point $p_i$ in frame $i$. To find the homography, we annotate four reference points (corners of the object) on the object in each frame. The natural environment prevents us from using a measurement arm \cite{lieberknecht2009dataset}, markers \cite{gauglitz2011evaluation} or SLAM system \cite{hare2012efficient} to obtain the ground truth. In \cite{roy2015tracking}, three tracking algorithms were used to annotate the ground truth. Despite still requiring manual verification as the final step, this approach is not suitable for the cases where the three algorithms fail to reach a correct consensus, especially for challenging scenarios. In this paper,  we use a semi-automatic approach to annotate the ground truth. In particular, we annotate every other frame for each sequence and the ground truth of 52,710 frames are produced in total.

\begin{figure}[!t]
\centering
\subfigure[Invisible (Map-2)]{
\includegraphics[width=0.475\linewidth]{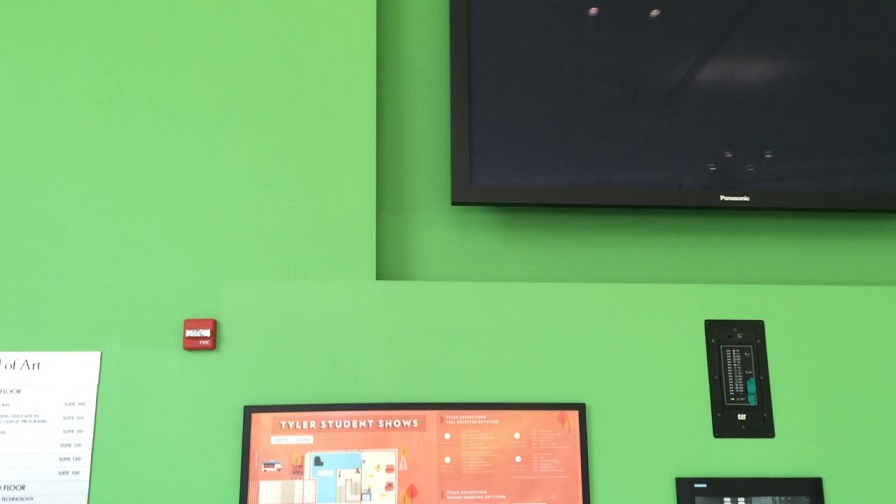}
}
\hspace{-8pt}
\subfigure[Blur (Painting-1)]{
\includegraphics[width=0.475\linewidth]{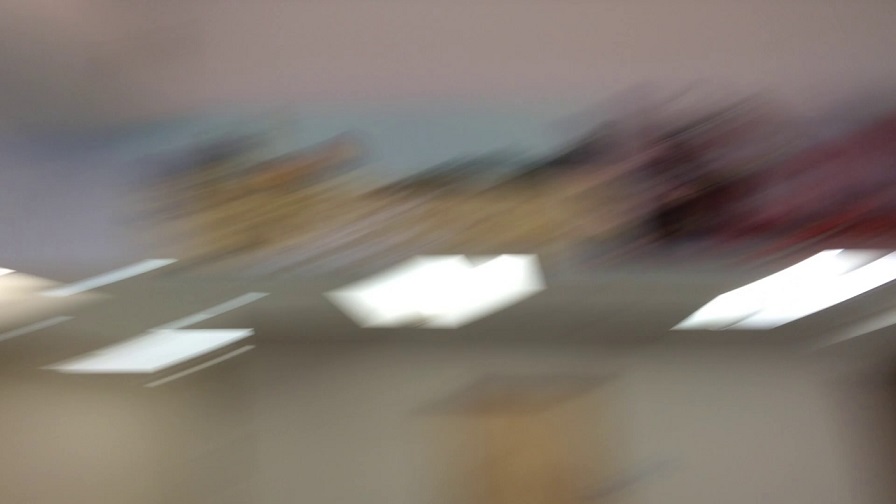}
}
\caption{Example frames excluded from annotation.}
\label{fig:exclusion}
\end{figure}

Fig.~\ref{fig:annotation-tool} shows the user interface of our annotation tool.  Besides the four corner points, we use four additional points located around the middle of the four edges to deal with occlusion and out-of-view.  On the top of Fig.~\ref{fig:annotation-tool} , it shows the initial eight points for reference; on the bottom, it is the current frame that needs annotation. The black margin around the image is used to help annotate the frames that are out-of-view. The annotation contains two steps:
\begin{itemize}
\item \textbf{Step 1:} Run the keypoint-based algorithm \cite{hare2012efficient} to get an initial estimation of the object state. Manual re-initialization of the algorithm is used so that the algorithm can better adapt to the change of the object state.
\item \textbf{Step 2:} Select four out of the eight reference points and manually fine tune their positions, then re-estimate the homography transformation with the selected points. The global shape of the object is also taken into account when it is occluded or out-of-view.
\end{itemize}
Note that in step 2: (1) the four corner points are selected first if they are visible in the image; (2) the initial four middle points might not remain at the middle after homography transformation, so when we use the middle points, we also take the context around the initial positions in the reference frame into consideration;  (3) we mark frames in which more than half of the target is invisible (occluded or out-of-view, Fig.~\ref{fig:exclusion}(a)) and frames that are heavily blurred (Fig.~\ref{fig:exclusion}(b)). Such marked frames will not be used for evaluation.

In general, after excluding frames the invisible part of which are more than half or heavily blurred as shown in Fig.~\ref{fig:exclusion}, the above annotation approach generates accurate ground truth with manageable amount of human labor.

\section{Evaluation}
\label{sec:evaluation}
\subsection{Selected trackers}
To study the performance of modern visual trackers for planar object tracking, we select \NUM representative algorithms from three different groups.

\vspace{1.5mm}\noindent\textbf{Keypoint-based planar tracking}:
\vspace{1.5mm}

\textbf{SIFT \cite{lowe2004distinctive}} and \textbf{SURF \cite{bay2008speeded}}: To evaluate the performance of SIFT and SURF for planar object tracking on our benchmark, we follow the traditional keypoint-based tracking pipeline and use OpenCV for implementation. These two trackers contain three steps: (1) keypoints detection; (2) keypoint matching via nearest neighbour search; and (3) homography estimation using RANSAC \cite{fischler1981random}.

\textbf{FERNS \cite{ozuysal2010fast}}: FERNS formulates the keypoints recognition problem in a naive Bayesian classification framework. The appearance of the image patch surrounding a keypoint is described by hundreds of simple binary features (ferns) depending on the intensities of two pixels, then the class posterior probabilities are estimated. By shifting the computation burden to the training stage as \cite{lepetit2006keypoint}, the classification of keypoints can be performed very fast.

\textbf{SOL \cite{hare2012efficient}:} Structured output learning (SOL)  is used to combine keypoints matching and transformation estimation in a unified framework. The adopted linear structured SVM algorithm allows the object model to adapt to a given environment quickly. To speed up the algorithm, the classifier is approximated with a set of binary vectors and the binary descriptor BRIEF \cite{calonder2010brief} is used for keypoint matching. The keypoints are extracted by FAST \cite{rosten2010faster}.  With  binary representation and Hamming distance similarity, matching can be performed extremely fast using bitwise operations.

\vspace{1.5mm}\noindent\textbf{Region-based planar tracking}:
\vspace{1.5mm}

\textbf{ESM \cite{benhimane2004real}:} The transformation parameters  in \cite{benhimane2004real} is estimated by minimizing the sum-of-squared-difference between a given template and the current image. To solve the optimization problem efficiently, efficient second-order minimization (ESM) is used to estimate the second order approximation of the cost function. Compared with the Newton method, ESM does not need to compute the Hessian and has a higher convergence rate.

\textbf{IC \cite{baker2004lucas}}: To avoid re-evaluating the Hessian in every iteration in the Lucas-Kanade image alignment algorithm \cite{lucas1981iterative}, the inverse compositional (IC) algorithm switches the role of the template and the image. The resulted optimization problem has a constant Hessian and can be pre-computed. The proof of equivalence between IC and  Lucas-Kanade is provided in \cite{baker2004lucas}. 

\textbf{SCV \cite{richa2011visual}:} Being invariant to non-linear illumination variation, the sum of conditional variance (SCV) is employed to measure the similarity between a given template and the current image in \cite{richa2011visual}. The SCV tracker can be viewed as an extension of ESM.

\textbf{GO-ESM \cite{goesm2017}}: As gradient orientations (GO) is robust to illumination changes, it is used in GO-ESM along with denoising techniques to model the appearance of the target. GO-ESM also generalizes ESM to multidimensional features.

\vspace{1.5mm}\noindent\textbf{Generic object tracking}:
\vspace{1.5mm}

\textbf{GPF \cite{kwon2014geometric}}: Using deterministic optimization to estimate the spatial transformation for template-based tracking can result in local optima. To overcome this limitation, the authors of \cite{kwon2014geometric} formulate the problem in a geometric particle filter (GPF) framework on a matrix Lie group.
GPF uses the combination of the incremental PCA model \cite{ross2008incremental} and normalized cross correlation (NCC) score to model the appearance of the target.

\textbf{IVT \cite{ross2008incremental}}: To deal with appearance change of the target, IVT uses an incremental PCA algorithm to update the appearance model which is a eigenbasis learned off-line. The algorithm estimates an affine transformation for each frame with particle filter.

\textbf{L1APG \cite{bao2012real}}: To solve the $\ell_{1}$ norm minimization problem efficiently of the sparse linear representation of target appearance and improve its robustness, L1APG uses a mixed norm and an efficient optimization method based on accelerated proximal gradient (APG) approach. L1APG also estimates an affine transformation for each frame.

Note that all these three generic tracking algorithms are template-based and they can be attributed to the region-based group.
For all above \NUM algorithms except SIFT \cite{lowe2004distinctive} and SURF \cite{bay2008speeded}, we use their available source codes. For ESM \cite{benhimane2004real}, IC \cite{baker2004lucas} and SCV \cite{richa2011visual}, we increase the number of maximum iterations for solving the optimization problem to 200; for other trackers, we use their default parameter setting. The original number of iterations used by GO-ESM \cite{goesm2017} is 200.

\begin{figure*}[!t]
\centering
\scriptsize
\begin{tabular}
{@{\hspace{.0mm}}c@{\hspace{0mm}} @{\hspace{0mm}}c@{\hspace{0mm}}
 @{\hspace{0mm}}c@{\hspace{0mm}} @{\hspace{0mm}}c@{\hspace{0mm}}}
\includegraphics[width=0.25\linewidth,height=0.17\linewidth]{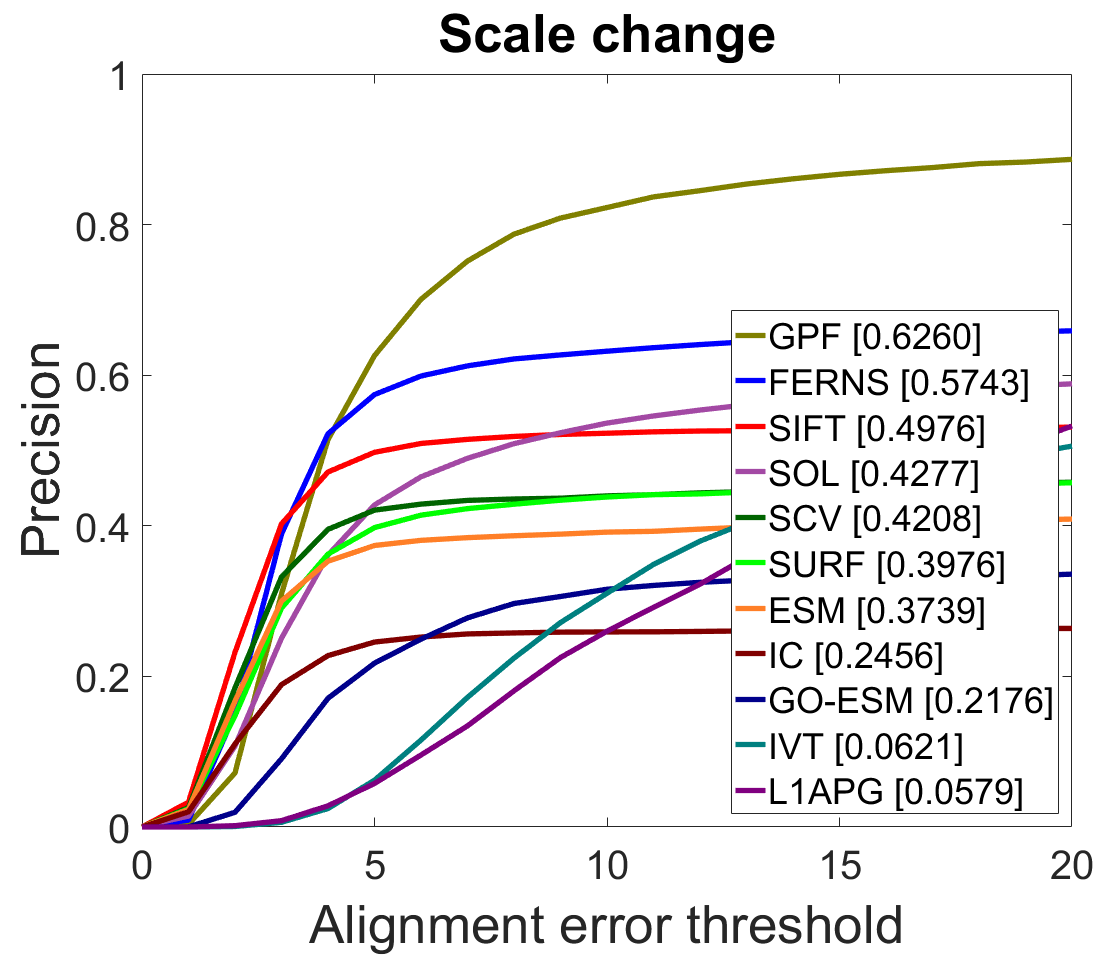}
&
\includegraphics[width=0.25\linewidth,height=0.17\linewidth]{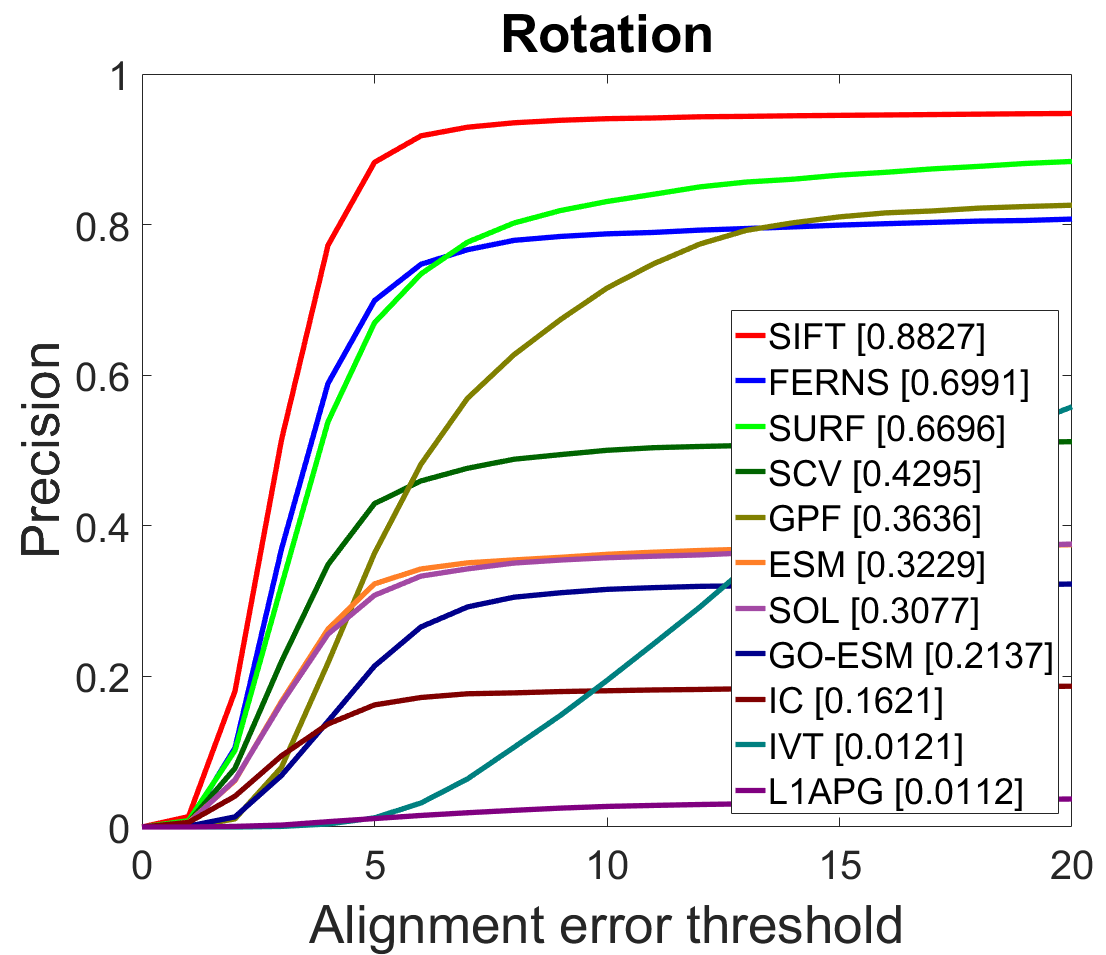}
&
\includegraphics[width=0.25\linewidth,height=0.17\linewidth]{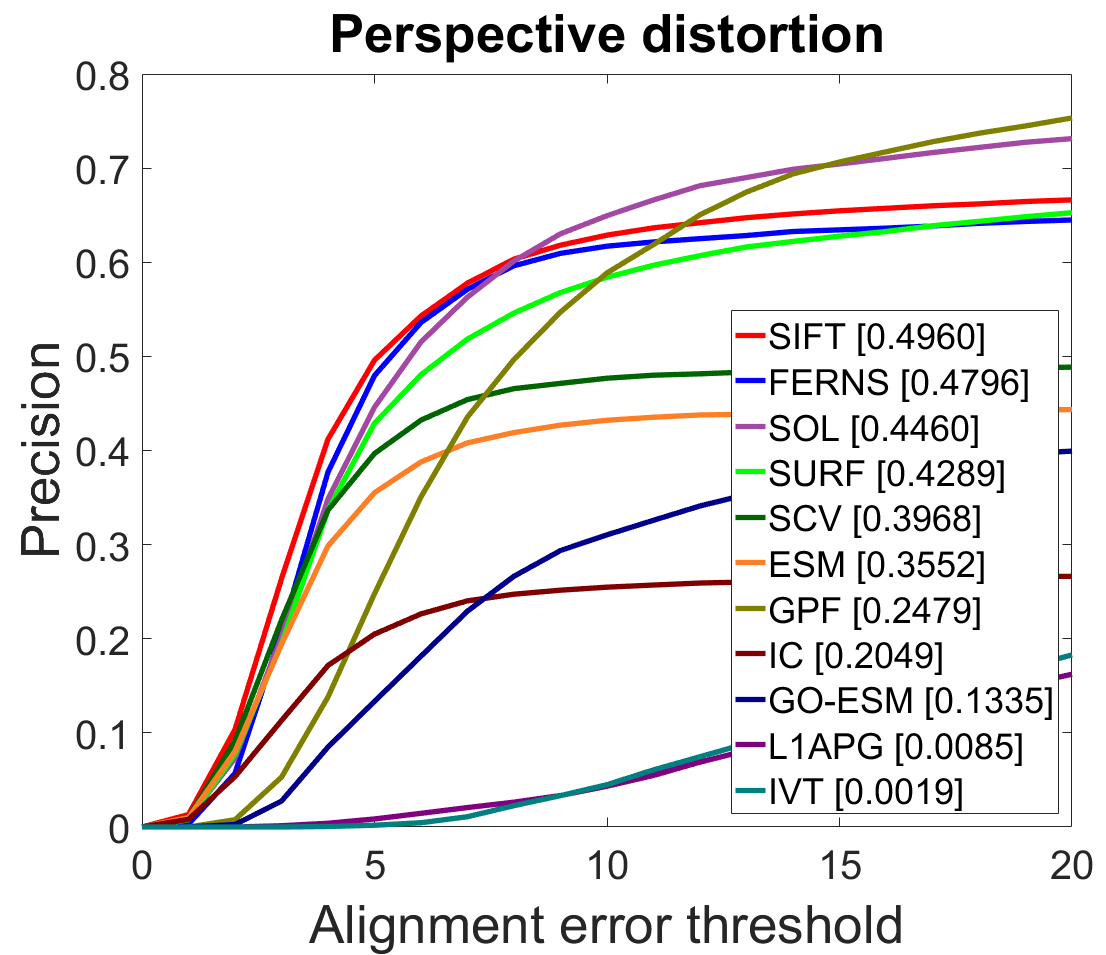}
&
\includegraphics[width=0.25\linewidth,height=0.17\linewidth]{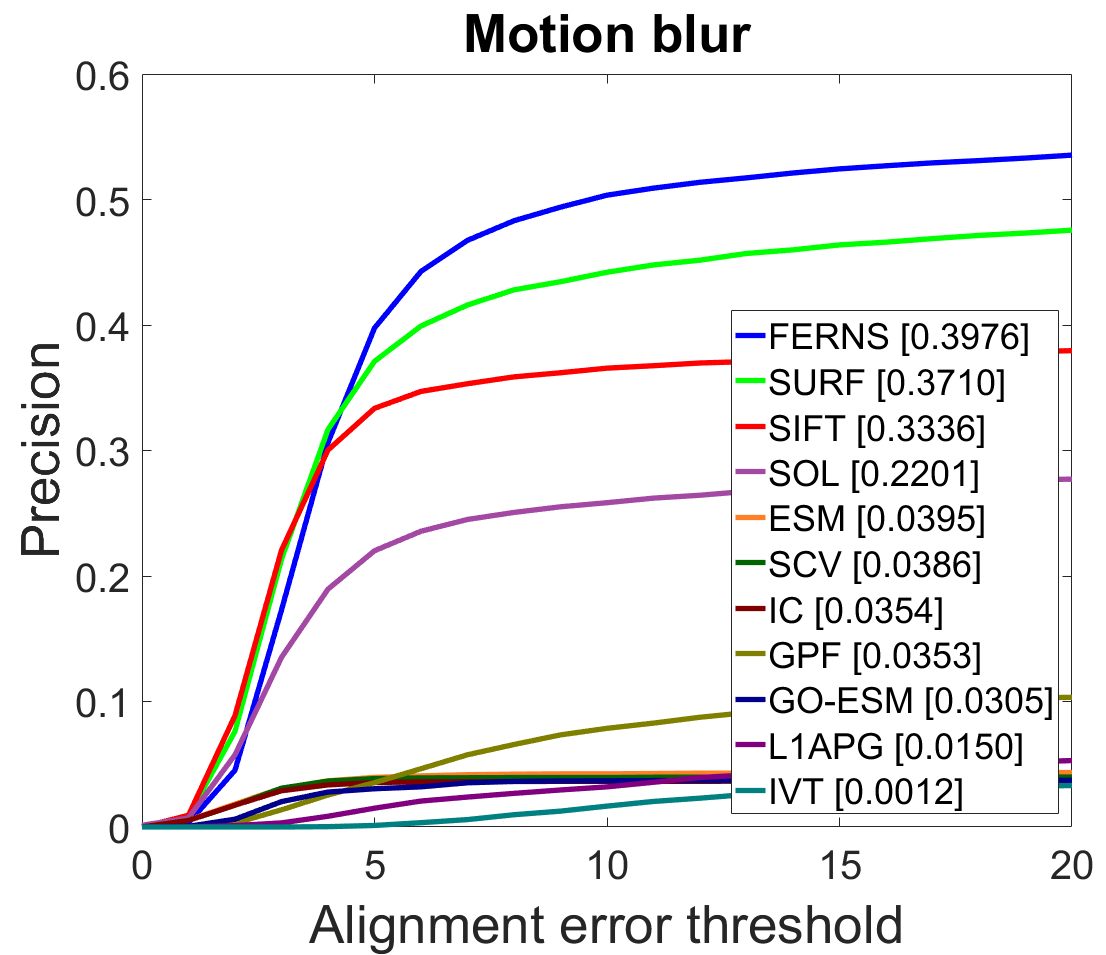}\\
(a)&(b)&(c)&(d)\\
\includegraphics[width=0.25\linewidth,height=0.17\linewidth]{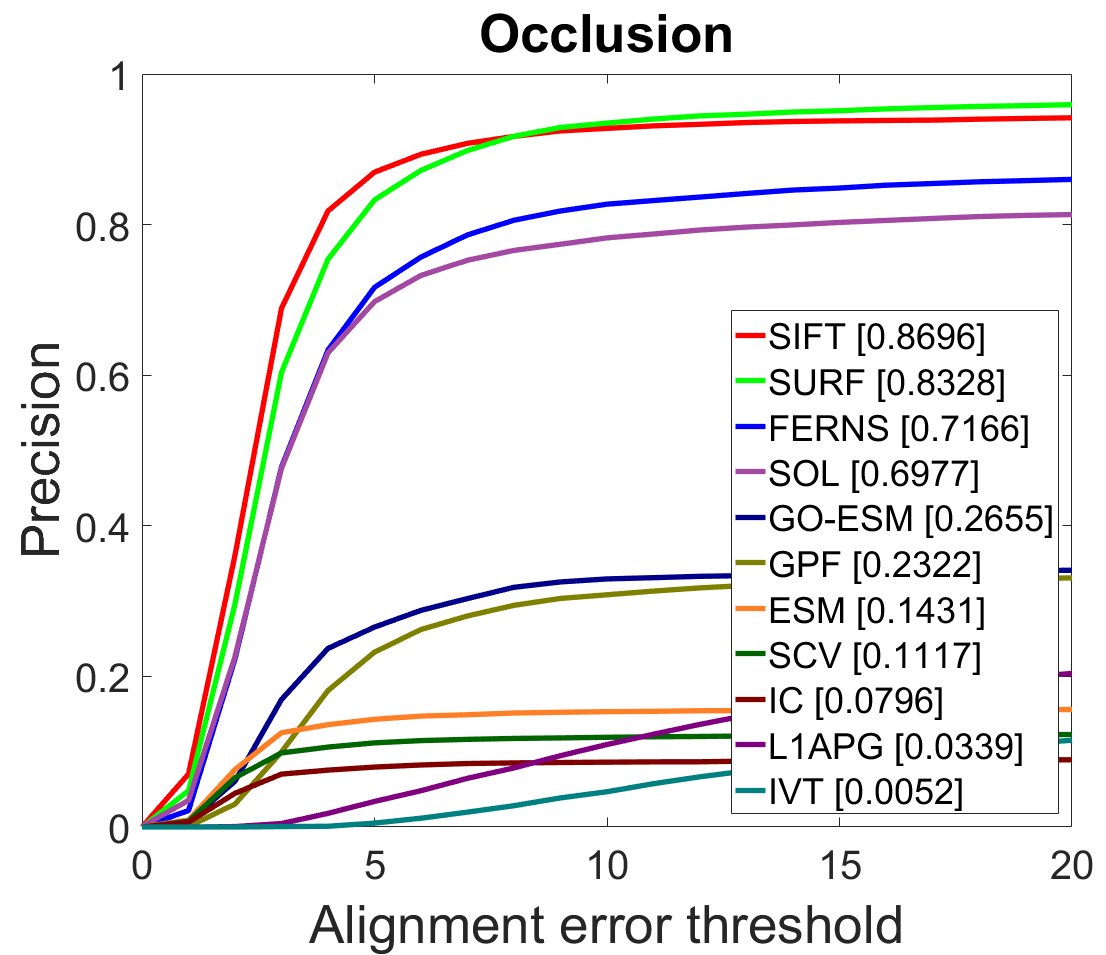}
&
\includegraphics[width=0.25\linewidth,height=0.17\linewidth]{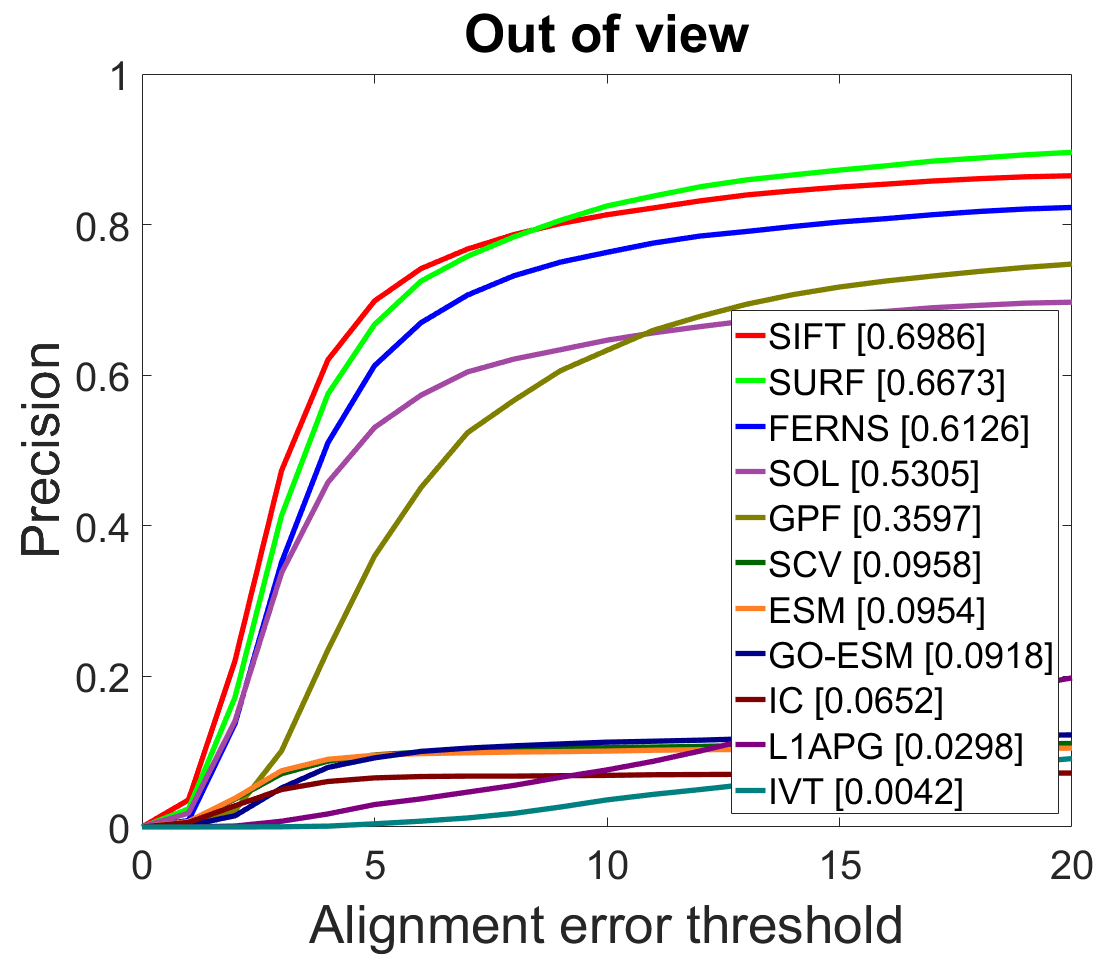}
&
\includegraphics[width=0.25\linewidth,height=0.17\linewidth]{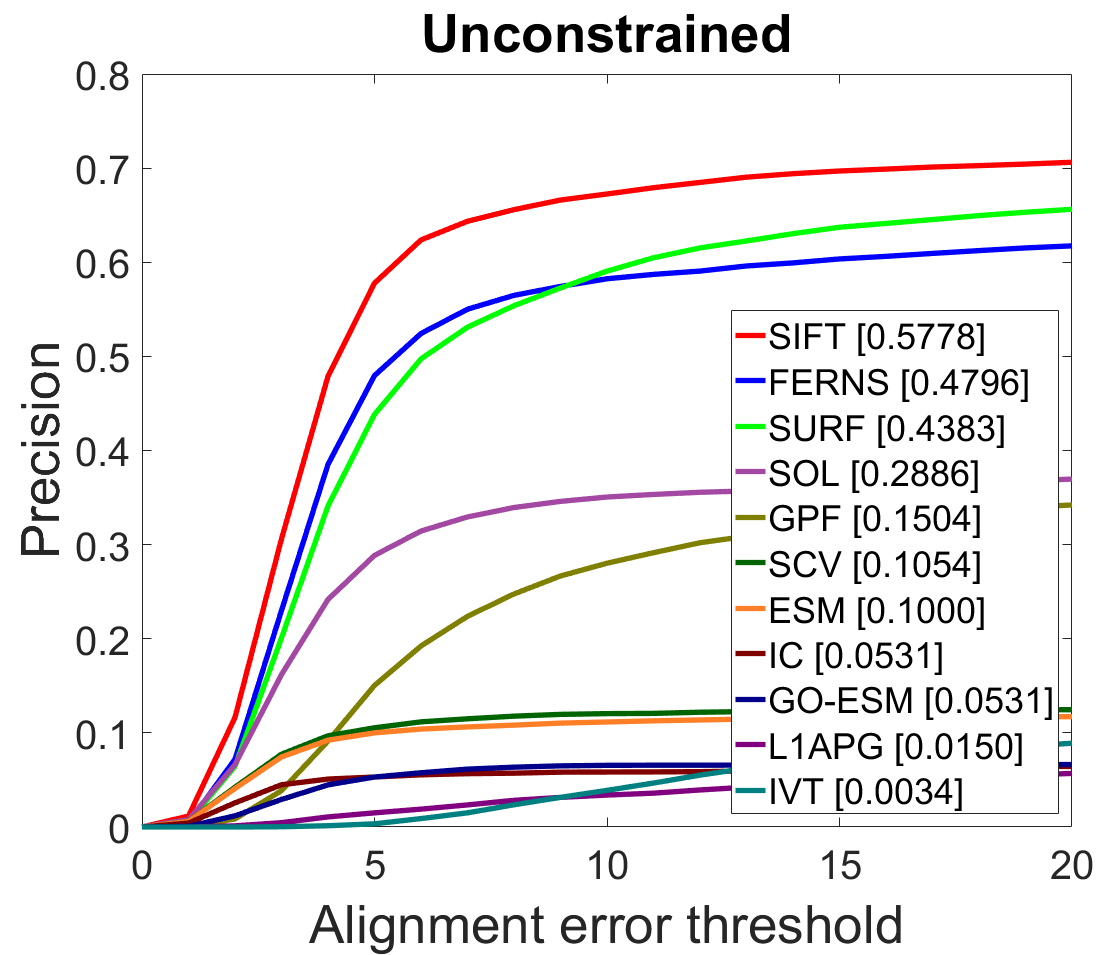}
&
\includegraphics[width=0.25\linewidth,height=0.17\linewidth]{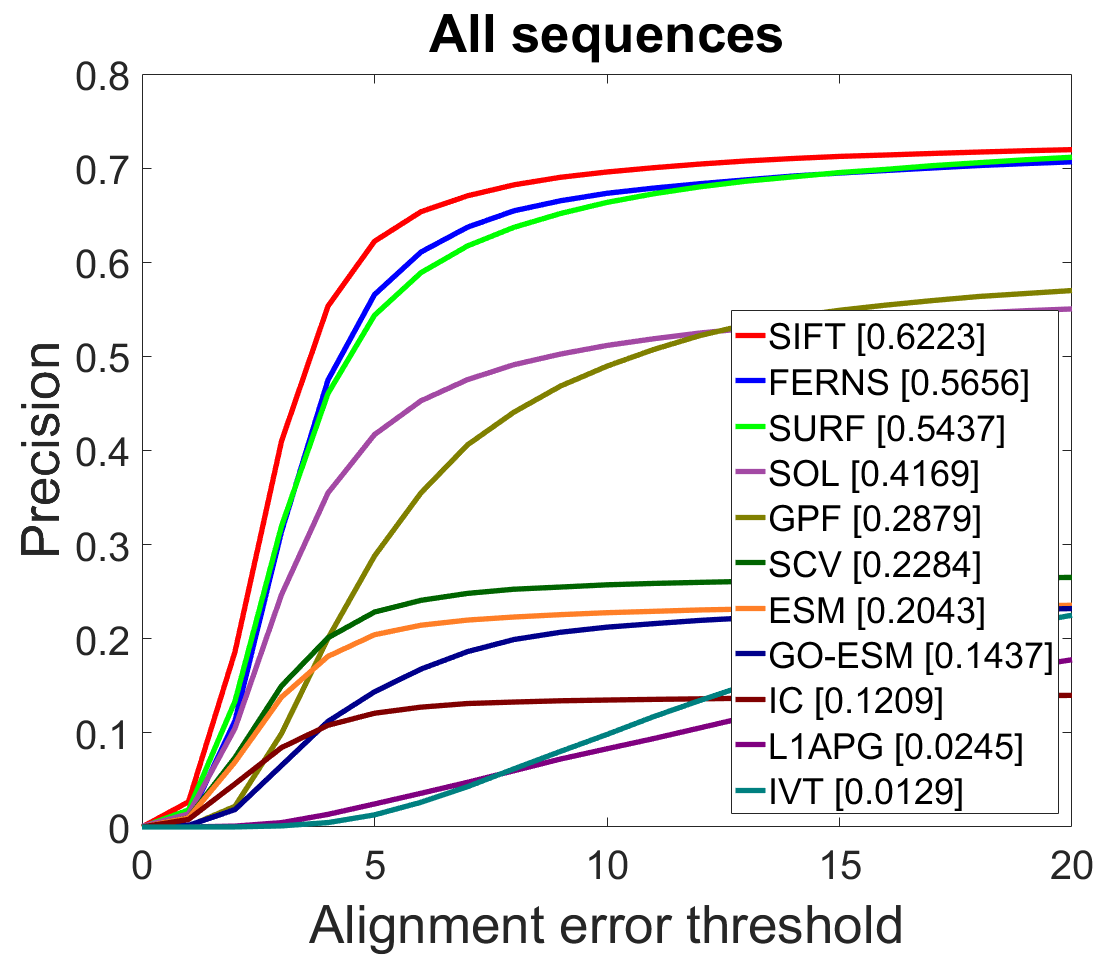}\\
(e)&(f)&(g)&(h)\\
\end{tabular}
\caption{Comparison of evaluated trackers using \emph{precision} plots. The precision at the threshold $t_p=5$ is used as a representative score.}
\label{fig:performance-comparison}
\end{figure*}

\begin{figure*}[!t]
\centering
\scriptsize
\begin{tabular}
{@{\hspace{.0mm}}c@{\hspace{0mm}} @{\hspace{0mm}}c@{\hspace{0mm}}
 @{\hspace{0mm}}c@{\hspace{0mm}} @{\hspace{0mm}}c@{\hspace{0mm}}}
\includegraphics[width=0.25\linewidth,height=0.17\linewidth]{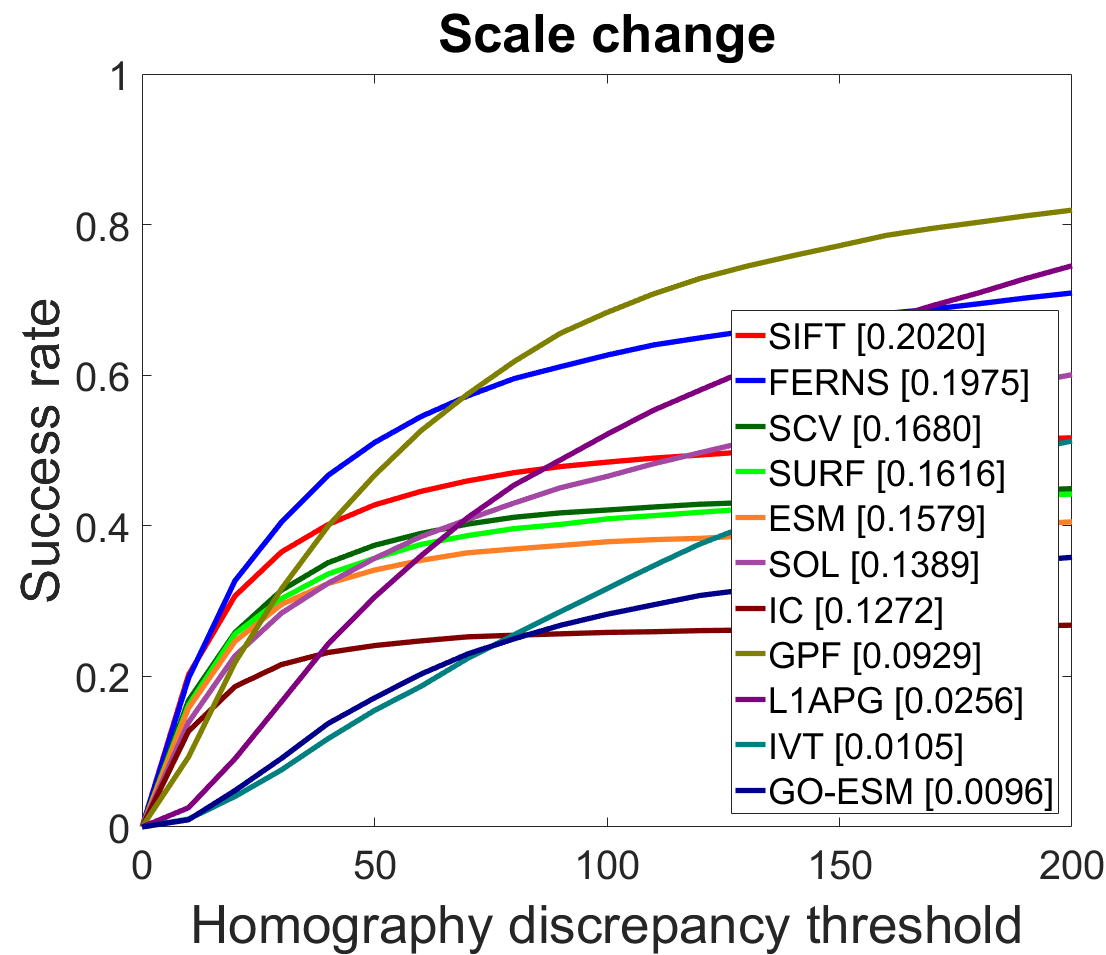}
&
\includegraphics[width=0.25\linewidth,height=0.17\linewidth]{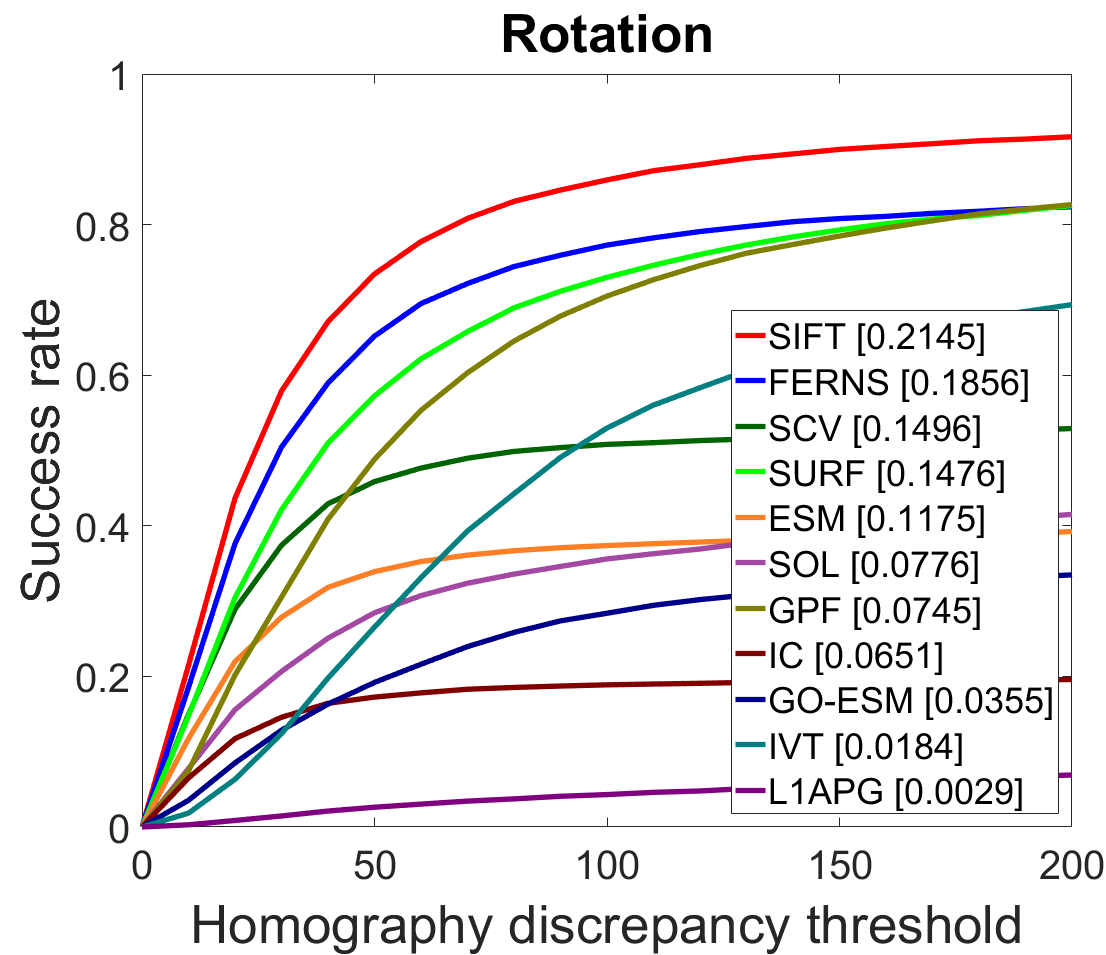}
&
\includegraphics[width=0.25\linewidth,height=0.17\linewidth]{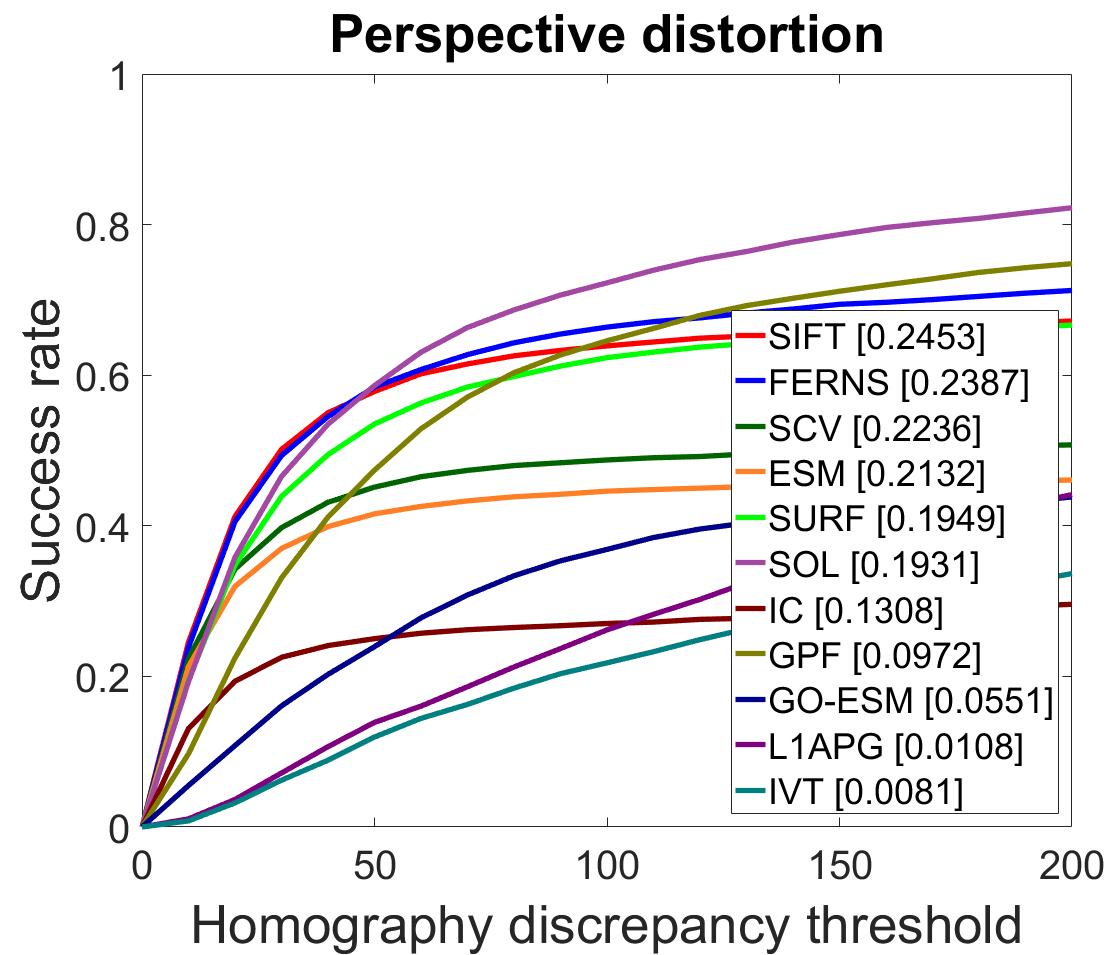}
&
\includegraphics[width=0.25\linewidth,height=0.17\linewidth]{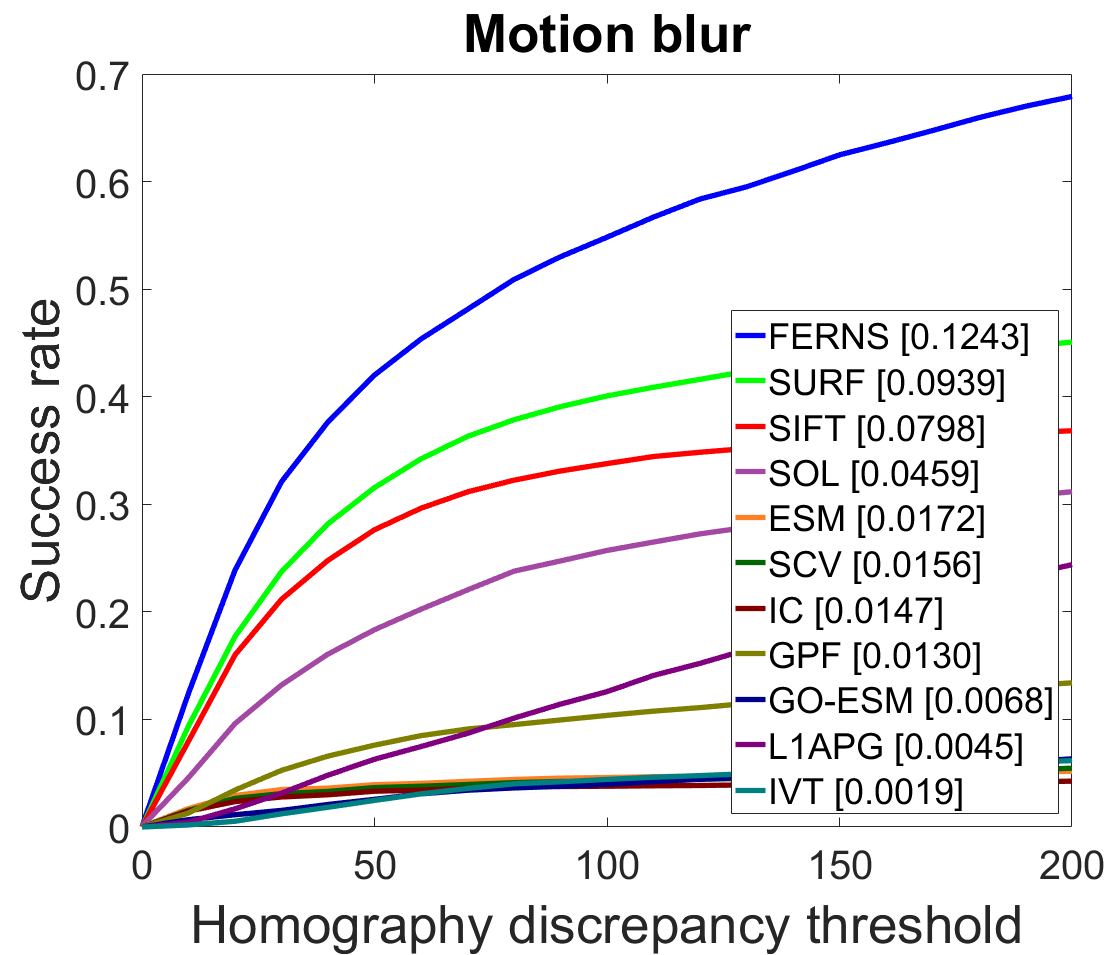}\\
(a)&(b)&(c)&(d)\\
\includegraphics[width=0.25\linewidth,height=0.17\linewidth]{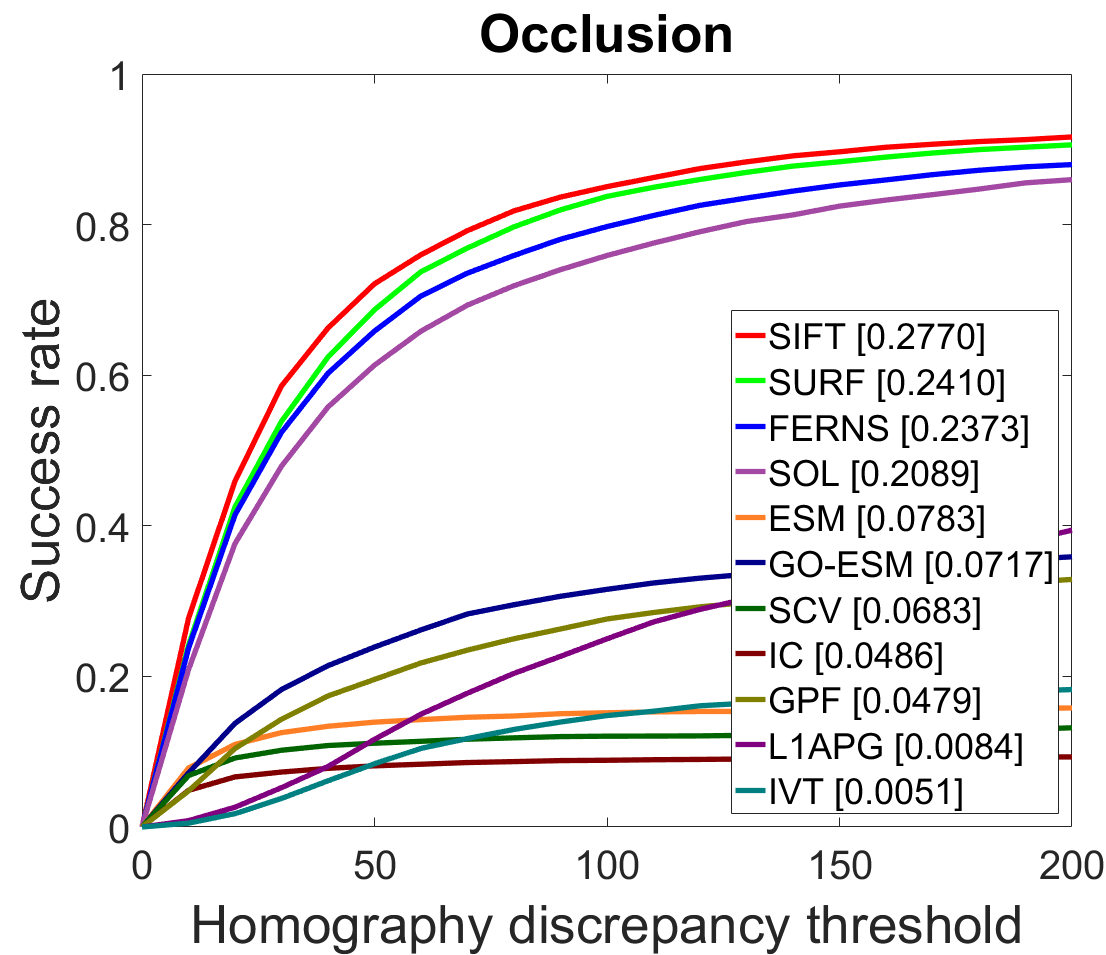}
&
\includegraphics[width=0.25\linewidth,height=0.17\linewidth]{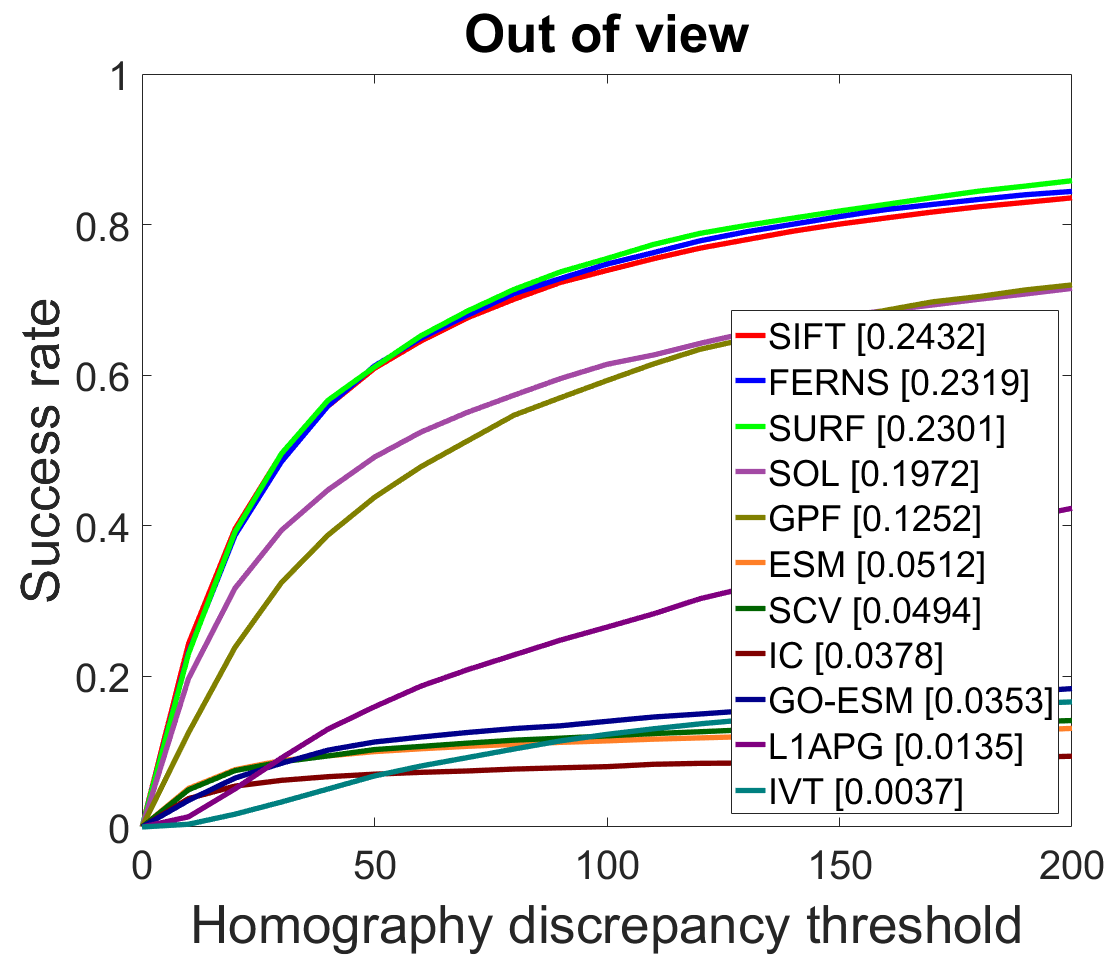}
&
\includegraphics[width=0.25\linewidth,height=0.17\linewidth]{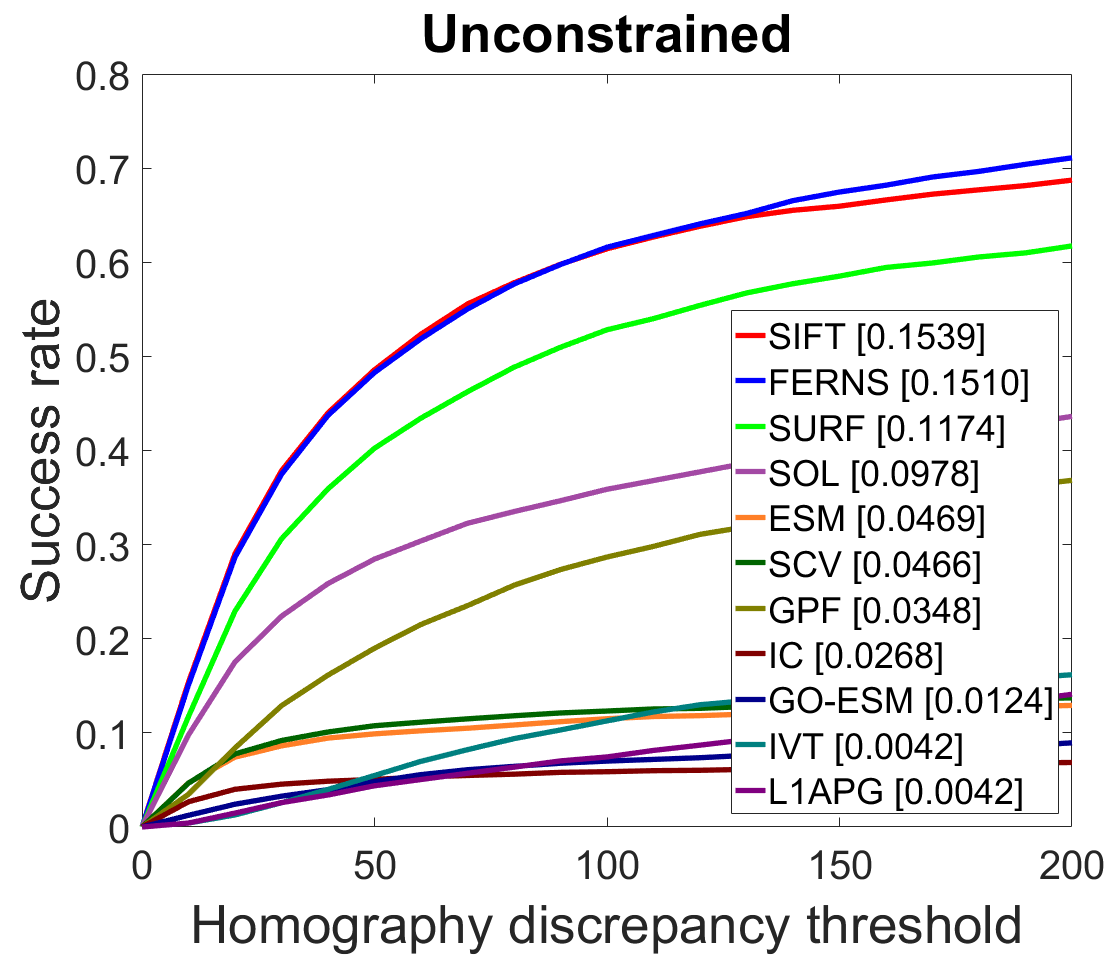}
&
\includegraphics[width=0.25\linewidth,height=0.17\linewidth]{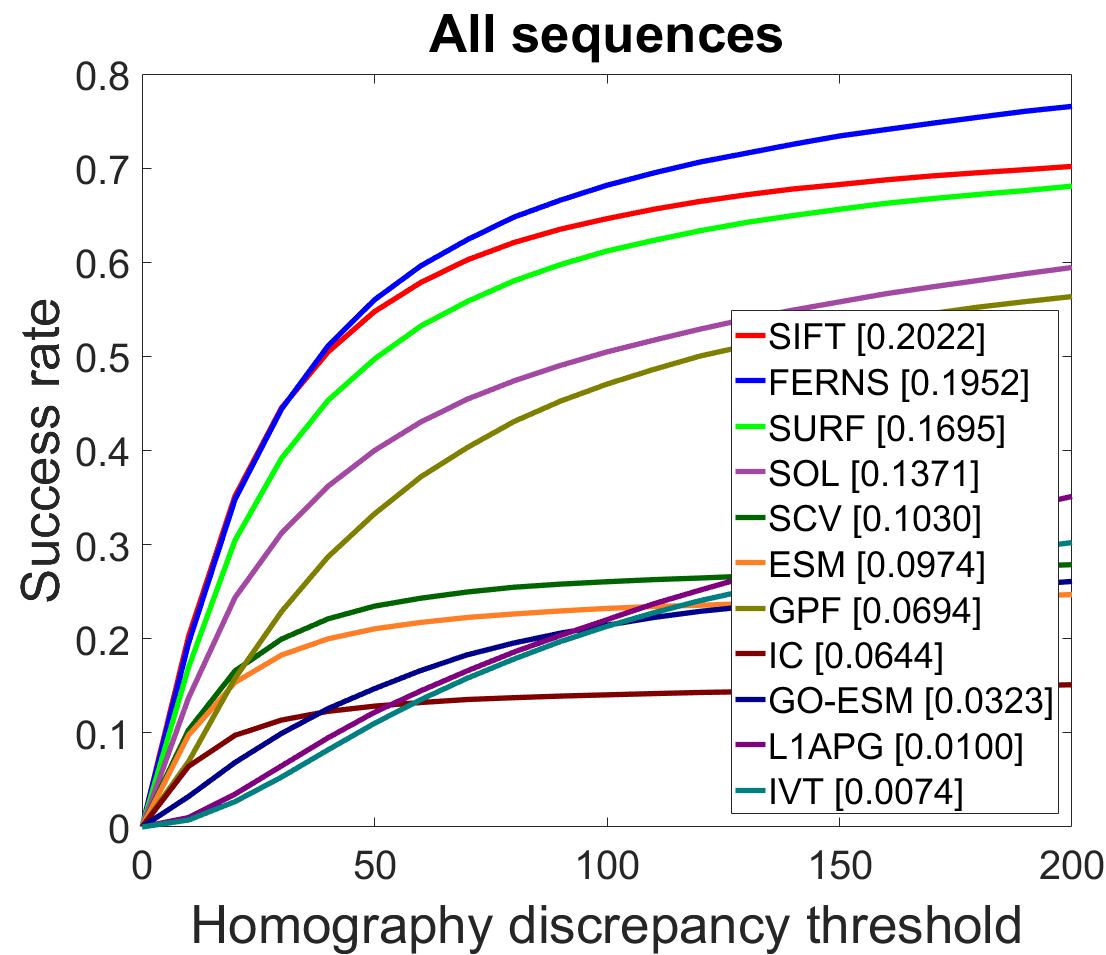}\\
(e)&(f)&(g)&(h)\\
\end{tabular}
\caption{Comparison of evaluated trackers using \emph{success} plots. The success rate at the threshold $t_s=10$ is used as a representative score.}
\label{fig:performance-comparison-homography}
\end{figure*}


\subsection{Evaluation metrics}\label{sec:metrics}
In this paper, we use the following two metrics to analyze the results quantitatively.

\noindent\textbf{Alignment error.}
The alignment error is based on the four reference points (four corners of the object), and is defined as the root of the mean square distances between the estimated positions of the points and their ground truth \cite{lieberknecht2009dataset,roy2015tracking},
\begin{equation}\label{eq:alignment-error}
  e_{AL} = \Big({\frac{1}{4}\sum_{i=1}^{4}(\mathbf{x}_i-\mathbf{x}_i^*)^2} \Big)^{1/2}
\end{equation}
where $\mathbf{x}_i$ is the position of a reference point and $\mathbf{x}_i^{*}$ is its ground truth position.

Precision plot has been adopted to evaluate the tracking algorithms for general purposes recently \cite{wu2015object}. In this work, we draw precision plot based on the alignment error, and it shows the percentage of frames whose $e_{AL}$ is smaller than a threshold $t_p$. We use $t_p=5$ as a representative precision score for each algorithm.

\noindent\textbf{Degree of difficulty of each object.} To rank the 30 planar objects used in our benchmark as shown in Fig.~\ref{fig:object-show}, we quantitatively derive the \emph{degree of difficulty} (DoD) of each object. Specifically, during the evaluation process, the precision score at the threshold $t_p=5$ for each sequence and each tracker is recorded. Then, given an object $obj$, its \emph{degree of difficulty} is defined as:
\begin{equation}
    \mathrm{DoD}_{obj} = 1 - (\mathrm{mean~precision~over~all~results~on~}obj). \nonumber
\end{equation}


\noindent\vspace{1.2mm}\textbf{Homography discrepancy.}
Homography discrepancy  measures the difference between the ground truth homography $T^*$ and the predicted one $T$, and it is defined as  \cite{hare2012efficient}:
\begin{equation}\label{eq:homography}
S(T^*,T)=\frac{1}{4}\sum_{i=1}^{4}\|\mathbf{c}_i-(T^*T^{-1})(\mathbf{c}_i)\|_2
\end{equation}
where $\{\mathbf{c}_i\}_{i=1}^{4}=\{(-1,-1)^{\top},(1,-1)^{\top},(-1,1)^{\top},(1,1)^{\top}\}$ are the corners of a square. $S(T^*,T)$ is $0$ if $T^*$ and $T$ are identical. The success rate of a tracker on a sequence is the percentage of frames whose homography discrepancy score is less than a threshold. We generate the success plot by varying the threshold from 0 to 200. Following \cite{hare2012efficient}, the success rate at threshold $t_s=10$ is used as a representative score.

\begin{figure}[!t]
\centering
\footnotesize\begin{tabular}
{@{\hspace{.0mm}}c@{\hspace{0.73mm}}@{\hspace{0.mm}}c@{\hspace{0.73mm}} @{\hspace{0.mm}}c@{\hspace{.0mm}}}
\includegraphics[width=0.33\linewidth]{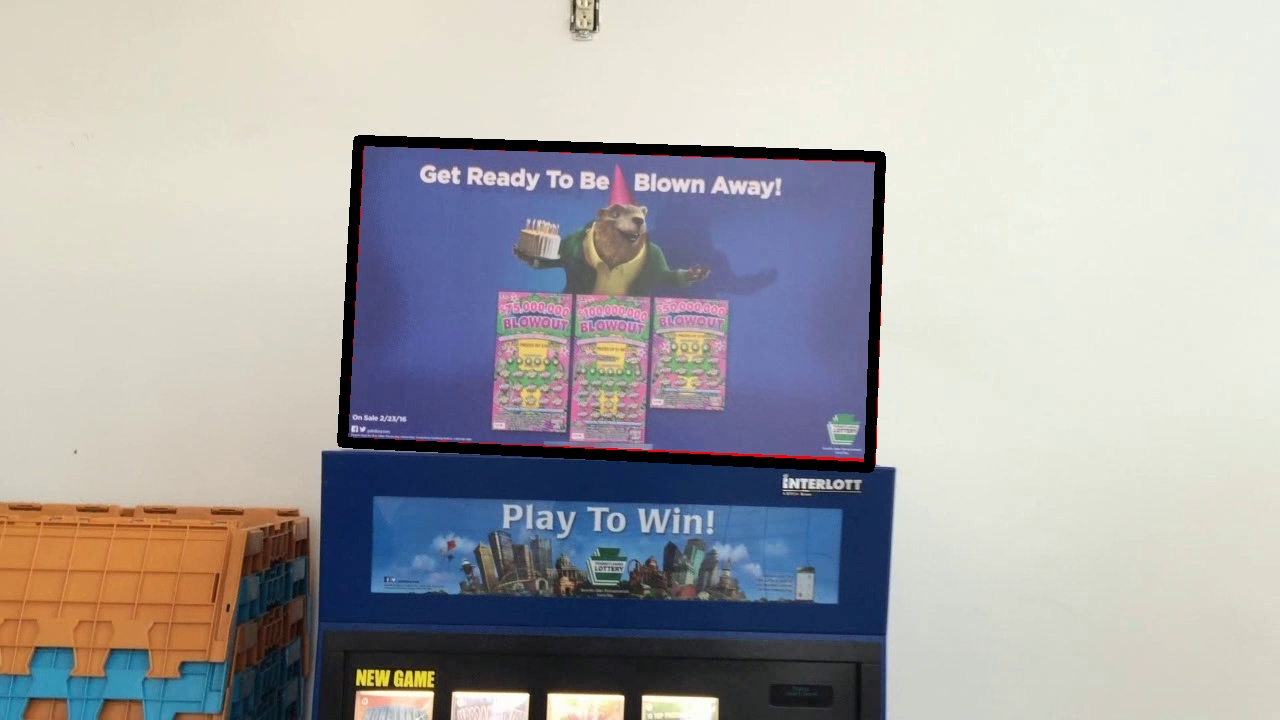}
&
\includegraphics[width=0.33\linewidth]{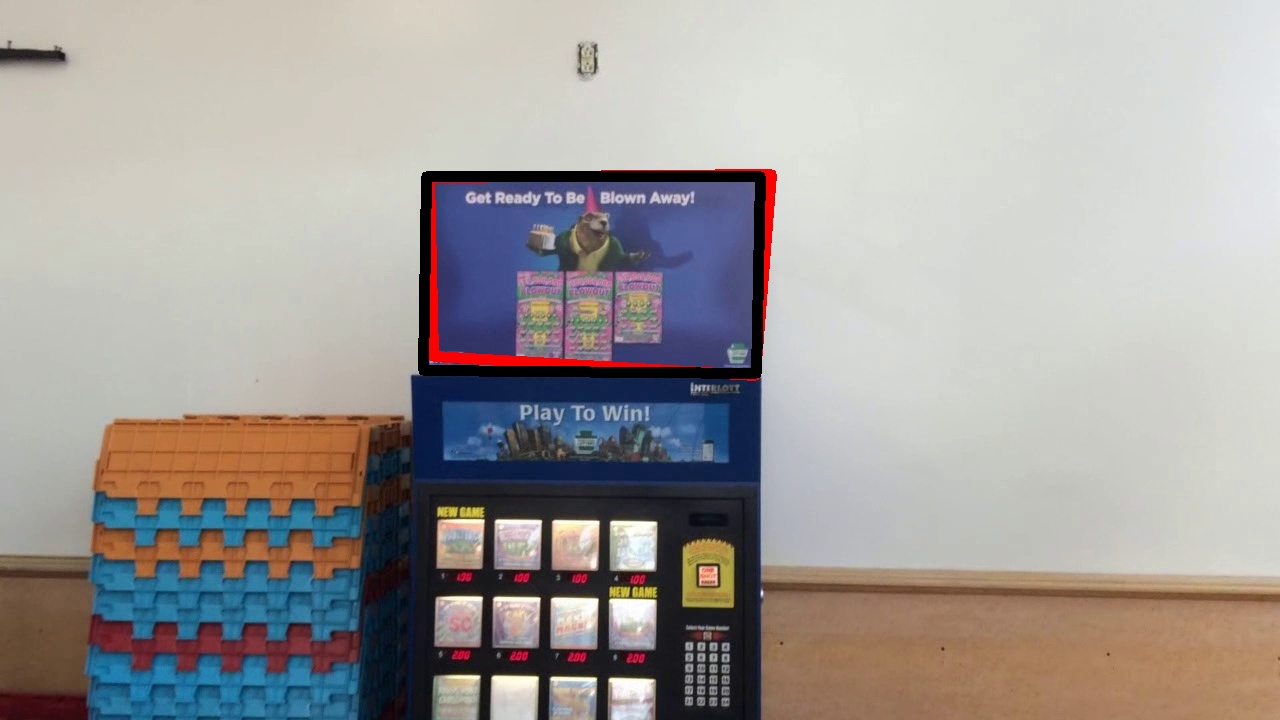}
&
\includegraphics[width=0.33\linewidth]{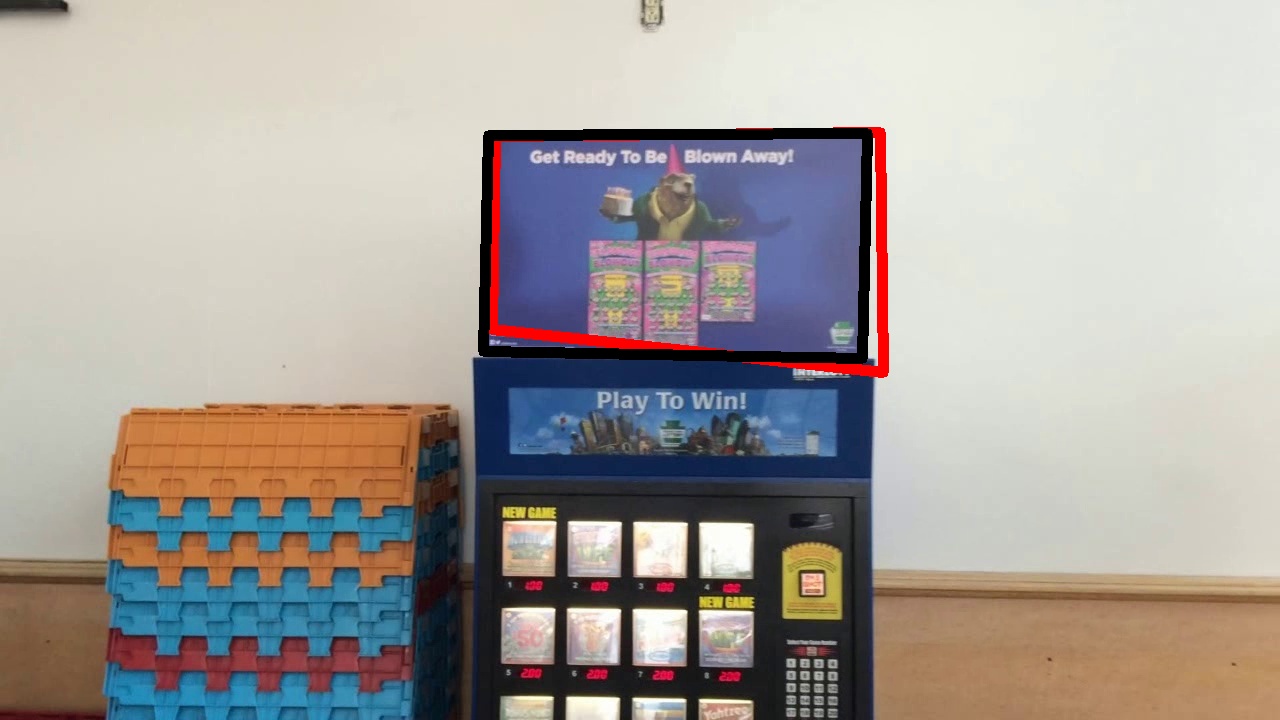}\\
(a) 8.05 &(b) 85.75 &(c) 315.75
\end{tabular}
\caption{Some example homography discrepancy scores (shown under subfigures). The black bounding box represents ground truth while the red one represents tracking result.}
\label{fig:show-homo-discrepancy}
\end{figure}

Note: (1) the same $t_s$ for success rate of different sequences may correspond to different $t_p$ for precision score;  (2) $t_s=10$ is a very tight threshold, as shown by some illustrative examples in Fig.~\ref{fig:show-homo-discrepancy}; and (3) as there is no correspondence between $t_s=10$ for success rate and $t_p=5$ for precision score, there are inconsistencies between the rank of trackers in Fig.~\ref{fig:performance-comparison} and Fig. \ref{fig:performance-comparison-homography}.

\subsection{Results and analysis}

\begin{figure*}[!t]
\centering
\subfigure[{\footnotesize Failure in scale change}]{
\label{fig:failure-sc}
    \includegraphics[width=0.158\linewidth,height=0.05\linewidth]{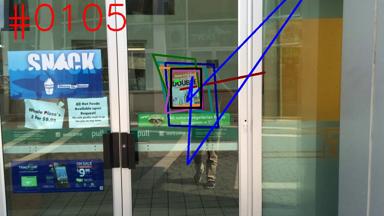}
    \hfill
    \includegraphics[width=0.158\linewidth,height=0.05\linewidth]{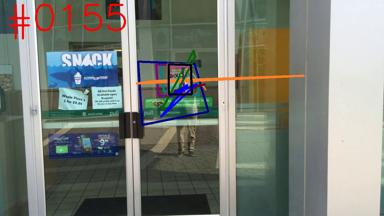}
    \hfill
    \includegraphics[width=0.158\linewidth,height=0.05\linewidth]{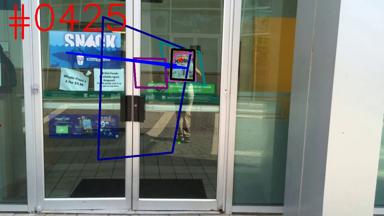}
}
\hspace{-10pt}
\subfigure[{\footnotesize Failure in rotation}]{
\label{fig:failure-rt}
    \includegraphics[width=0.158\linewidth,height=0.05\linewidth]{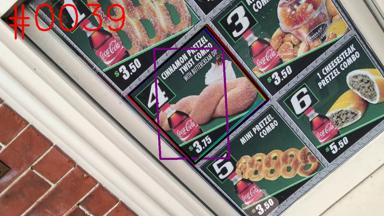}
    \hfill
    \includegraphics[width=0.158\linewidth,height=0.05\linewidth]{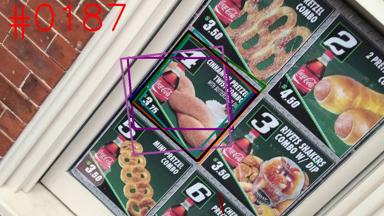}
    \hfill
    \includegraphics[width=0.158\linewidth,height=0.05\linewidth]{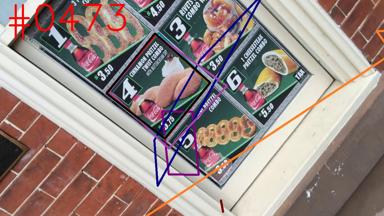}
}
\\\vspace{-.7mm}
\vspace{-3pt}\subfigure[{\footnotesize Failure in perspective distortion}]{
\label{fig:failure-pd}
    \includegraphics[width=0.158\linewidth,height=0.05\linewidth]{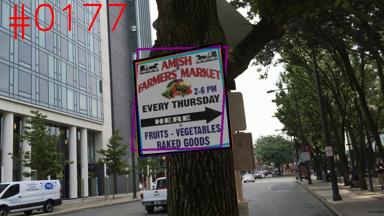}
    \hfill
    \includegraphics[width=0.158\linewidth,height=0.05\linewidth]{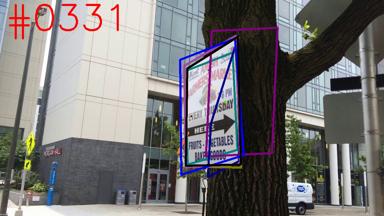}
    \hfill
    \includegraphics[width=0.158\linewidth,height=0.05\linewidth]{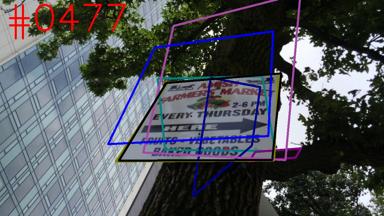}
}
\hspace{-10pt}
\subfigure[{\footnotesize Failure in motion blur}]{
\label{fig:failure-mb}
    \includegraphics[width=0.158\linewidth,height=0.05\linewidth]{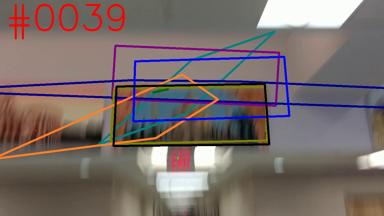}
    \hfill
    \includegraphics[width=0.158\linewidth,height=0.05\linewidth]{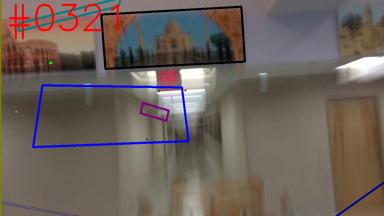}
    \hfill
    \includegraphics[width=0.158\linewidth,height=0.05\linewidth]{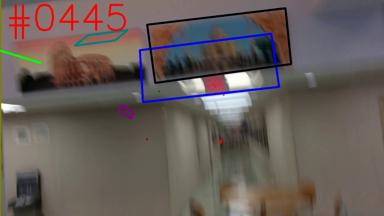}
}
\\\vspace{-.7mm}
\vspace{-3pt}\subfigure[{\footnotesize Failure in occlusion}]{
\label{fig:failure-occ}
   \includegraphics[width=0.158\linewidth,height=0.05\linewidth]{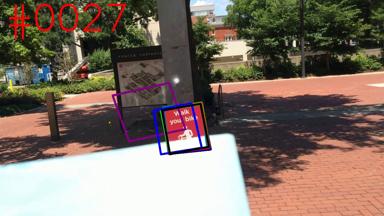}
    \hfill
    \includegraphics[width=0.158\linewidth,height=0.05\linewidth]{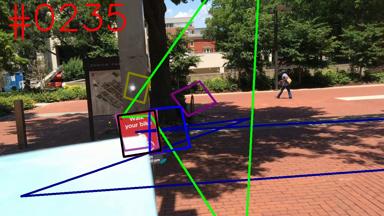}
    \hfill
    \includegraphics[width=0.158\linewidth,height=0.05\linewidth]{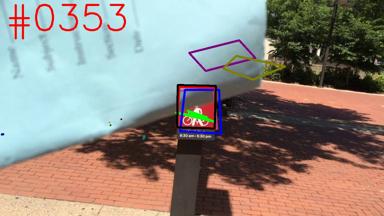}
}
\hspace{-10pt}
\subfigure[{\footnotesize Failure in out-of-view}]{
\label{fig:failure-ov}
    \includegraphics[width=0.158\linewidth,height=0.05\linewidth]{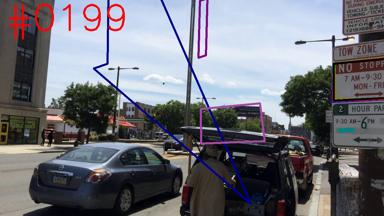}
    \hfill
    \includegraphics[width=0.158\linewidth,height=0.05\linewidth]{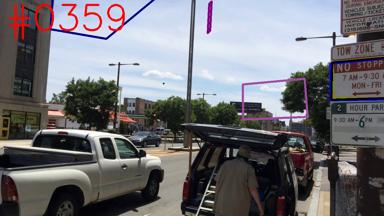}
    \hfill
    \includegraphics[width=0.158\linewidth,height=0.05\linewidth]{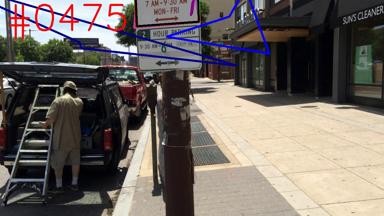}
}
\\\vspace{-.9mm}
\subfigure[{\footnotesize Failure in unconstrained}]{
\label{fig:failure-occ}
    \includegraphics[width=0.158\linewidth,height=0.05\linewidth]{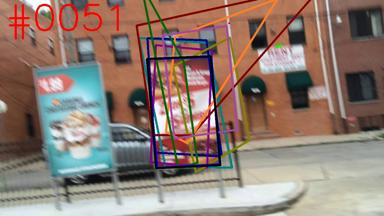}
    \hfill
    \includegraphics[width=0.158\linewidth,height=0.05\linewidth]{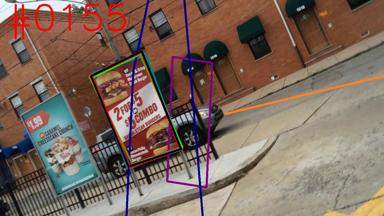}
    \hfill
    \includegraphics[width=0.158\linewidth,height=0.05\linewidth]{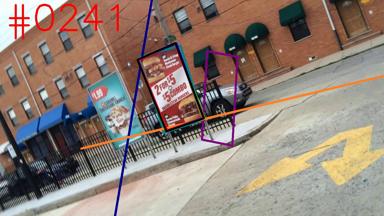}
    \hfill
    \includegraphics[width=0.158\linewidth,height=0.05\linewidth]{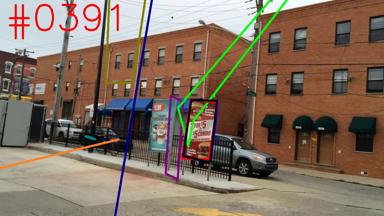}
    \hfill
    \includegraphics[width=0.158\linewidth,height=0.05\linewidth]{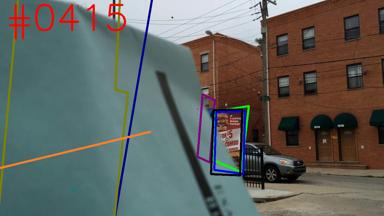}
    \hfill
    \includegraphics[width=0.158\linewidth,height=0.05\linewidth]{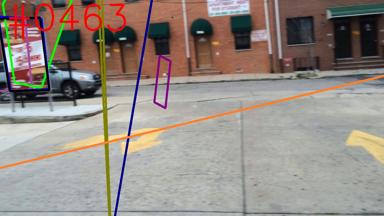}
}
\\\vspace{-.7mm}
\includegraphics[width=0.7\linewidth]{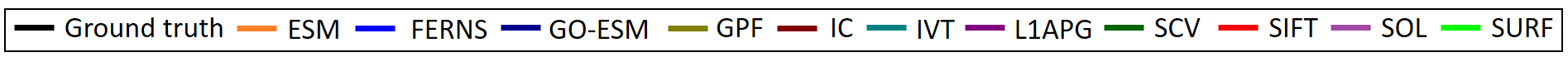}
\caption{Some failures observed in our experiment involving different challenge factors.}
\label{fig:failure}
\end{figure*}

\subsubsection{Comparison with respect to different challenges}

Fig.~\ref{fig:performance-comparison} shows the comparison among the \NUM trackers by precision plot using both subsets of sequences according to different motion patterns and all the sequences. In addition, the success plots of each tracker using the homography discrepancy are reported in Fig. \ref{fig:performance-comparison-homography}. It is worth noting that the performance of the generic object trackers IVT \cite{ross2008incremental} and L1APG \cite{bao2012real} are obviously worse than other trackers. One possible reason is that the parameters of these two trackers are set for the tracking scenario which just requires coarse bounding box estimation; another possible reason is that the adopted affine transformation with six degree-of-freedom is not sufficient to get very accurate results. In the following part, we use the performances of the other nine trackers for analysis purpose. Also, as the alignment error is more easy to measure perceptually and the success rate at $t_s=10$ is too tight (as shown in Fig.~\ref{fig:show-homo-discrepancy}),  we mainly use the precision plots to analyze the results and success plots are displayed for further validation purpose.

\begin{figure}[!t]
\centering
\begin{tabular}
{@{\hspace{.0mm}}c@{\hspace{.0mm}} @{\hspace{0.mm}}c@{\hspace{.0mm}}}
\includegraphics[width=0.5\linewidth]{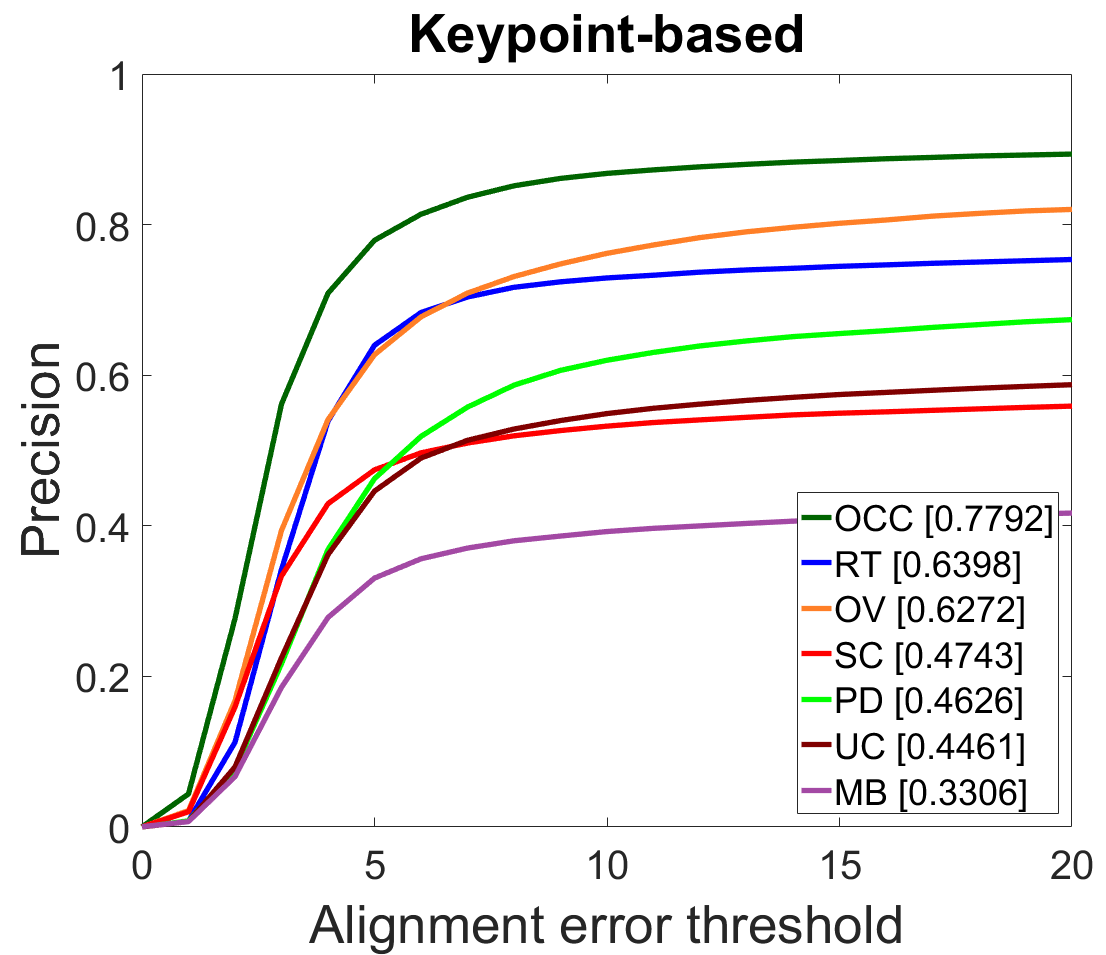}
&
\includegraphics[width=0.5\linewidth]{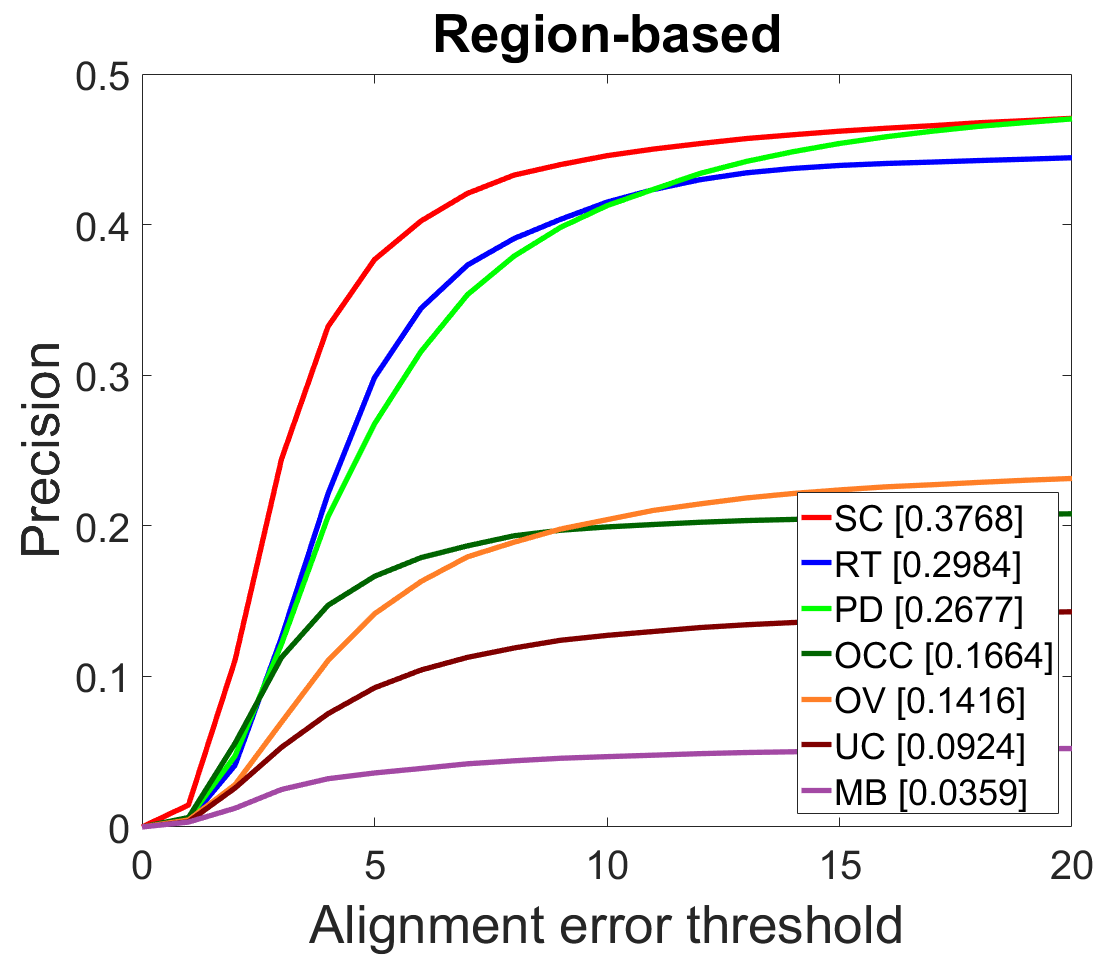}\\
(a)&(b)
\end{tabular}
\caption{The overall performance of trackers in two groups for different challenging factors. For each group, the overall performance is calculated by averaging the performances of trackers within this group. The precision at the threshold $t_p=5$ is used.
}
\label{fig:performance-precision}
\end{figure}

For scale change (Fig.~\ref{fig:performance-comparison}(a)), GPF performs best and FERNS also achieves comparable performance. Though SURF is also designed to be scale invariant, its performance is not promising on this subset. For the rotation subset (Fig.~\ref{fig:performance-comparison}(b)), although all of SIFT, FERNS and SURF are designed to be rotation-invariant, SIFT outperforms the other two algorithms by a large margin. Also, SCV and GPF achieves better results than other region-based trackers. The relatively inferior performance of SOL should be due to that the BRIEF descriptor lacks invariance ability to in-plane rotation \cite{calonder2010brief}.

Under the perspective distortion subset (Fig.~\ref{fig:performance-comparison}(c)), all the keypoint-based trackers outperform all the region-based trackers. The performances of SIFT, FERNS and GPF decrease obviously compared with scale change or rotation. SIFT itself is not designed to be invariant to perspective distortion. During the training stage of FERNS, it generates training samples with randomly picked affine transformation, nevertheless, the perspective distortion can also be homography transformation. For SCV and ESM, they have similar performance across these three motion patterns.

Motion blur (Fig.~\ref{fig:performance-comparison}(d)) is the most challenging motion pattern for all these \NUM trackers. As motion blur deteriorates the quality of the entire image, it is difficult for keypoint-based trackers to detect useful keypoints, and for region-based trackers to measure the similarity between images patches effectively.

For occlusion (Fig.~\ref{fig:performance-comparison}(e)) and out of view (Fig.~\ref{fig:performance-comparison}(f)),  the performances of the keypoint-based trackers are obviously better than the region-based trackers. This is consistent with the fact that it is still possible to obtain a set of correspondences between the target and image keypoints when occlusion appears or the target is out-of-view, and the correspondences are accurate enough to estimate the geometric transformation correctly. However, for region-based based trackers, both occlusion and out of view can cause large appearance variance.


According to the performances with respect to the unconstrained subset of sequences (Fig.~\ref{fig:performance-comparison}(g)) and all the sequences (Fig.~\ref{fig:performance-comparison}(h)), in general, the keypoint-based trackers are more robust than the region-based trackers. The obvious performance difference can be attributed to the following two reasons: (1) though the image similarity measure SCV adopted by \cite{richa2011visual} or GO adopted by \cite{goesm2017} are robust to illumination variations, their robustness is not comparable with the state-of-art keypoint detectors and descriptors or ferns; and (2)  the keypoint-based algorithms  use the tracking-by-detection strategy and the detection in the current frame depends little on the object location in previous frames; by contrast,  the region-based algorithms make use of the previous object state to reduce the optimization space for efficiency. Thus it is easier for keypoint-based trackers to recover from failure than region-based trackers.

Also, for ESM based algorithms \cite{benhimane2004real,richa2011visual,goesm2017}, SCV \cite{richa2011visual} is a little better than the original ESM tracker \cite{benhimane2004real} using the sum-of-squared-difference for appearance similarity measure. Though gradient orientations is robust to illumination change, the overall performance of GO-ESM \cite{goesm2017} is worse than ESM \cite{benhimane2004real}.  At the same time, ESM, SCV and GO-ESM perform better than IC \cite{baker2004lucas}, implying that the efficient second-order minimization approach is better than the inverse compositional optimization approach for the planar object tracking task. Some failure cases based on different motion patterns are shown in Fig.~\ref{fig:failure}.


\subsubsection{Overall performance of trackers in each group} We summarize the overall performance of trackers in each group by average precision plot in Fig.~\ref{fig:performance-precision}(a) and Fig.~\ref{fig:performance-precision}(b) respectively. Note that we include the GPF tracker \cite{kwon2014geometric} in the region-based group, and we do not consider IVT \cite{ross2008incremental} and L1APG \cite{bao2012real} for these two figures. We rank the performance with respect to different challenging factors using the precision score at the threshold $t_p=5$.  

The average precision plot of the keypoint-based trackers \cite{lowe2004distinctive,hare2012efficient,ozuysal2010fast,bay2008speeded} in Fig.~\ref{fig:performance-precision}(a) shows that they are more robust to occlusion, rotation and out-of-view than to other challenging factors. This  is consistent with the better performance of keypoint-based trackers on these three subsets as shown in Fig.~\ref{fig:performance-comparison}(e), Fig.~\ref{fig:performance-comparison}(b) and Fig.~\ref{fig:performance-comparison}(f) respectively.
The most challenging situation for the keypoint-based trackers is motion blur, as motion blur heavily affects the repeatability of the keypoints and the associated appearance descritpion. 

The average precision plot of the region-based trackers \cite{richa2011visual,benhimane2004real,baker2004lucas,goesm2017,kwon2014geometric} is given in Fig.~\ref{fig:performance-precision}(b). It shows that the region-based trackers are more robust to scale change, rotation and perspective distortion than to occlusion and out-of-view. This observation is consistent with the fact that the region-based trackers find the transformation by directly minimizing the error that measures the similarity between the entire template and the image, and occlusion and out-of-view increase the dissimilarity largely between the template and the corresponding image patch after alignment. Motion blur remains the most challenging factor due to appearance corruption and large displacement of the target. 

\section{Conclusion}
\label{sec:conclusion}
In this paper, we present a benchmark for evaluating planar object tracking algorithms in the wild. The dataset is constructed according to seven different challenging factors so that the performance of trackers can be investigated thoroughly. We design a semi-manual approach to annotate the ground truth accurately. We also evaluate \NUM state-of-the-art algorithms on the dataset with two metrics and give detailed analysis. The evaluation result shows that there is large space for improvement for all algorithms. We expect that our work can provide dataset and motivation for future study on planar object tracking in unconstrained natural environments.

\section*{ACKNOWLEDGMENT}

This work is supported by China National Key Research and Development Plan (Grant No. 2016YFB1001200).


\end{document}